%% file: neurips_2026.tex
\documentclass{article}

\usepackage[main, final]{neurips_2026}

\usepackage[utf8]{inputenc} 
\usepackage[T1]{fontenc}    
\usepackage{hyperref}       
\usepackage{url}            
\usepackage{booktabs}       
\usepackage{amsfonts}       
\usepackage{nicefrac}       
\usepackage{microtype}      
\usepackage{xcolor}         
\usepackage{amsmath}
\usepackage{multirow}
\usepackage{multicol}
\usepackage{graphicx}
\usepackage{booktabs}
\usepackage{diagbox}
\usepackage[table]{xcolor}
\usepackage{tikz}
\usepackage{xspace}
\usepackage[normalem]{ulem}
\usepackage{subcaption}
\usepackage{algorithm}
\usepackage{algpseudocode}
\usepackage{amsmath,amssymb}
\usepackage{enumitem}

\usepackage{booktabs}
\usepackage{makecell}
\usepackage{graphicx}

\newcommand{\methodname}{AUV\xspace}

\title{Attack-Inference Aligned Universal Adversarial Attacks on Video Object Segmentation}

\definecolor{defaultColor}{RGB}{230, 230, 250}

\newcommand{\pub}[1]{\color{gray}{\scriptsize{[{#1}]}}}

\newcommand{\note}[1]{}

\author{%
  Ruiqi Shen$^{1}$ \quad Chang Liu$^{2\ *}$ \quad Henghui Ding$^{1}$\thanks{Corresponding Author.} \\
  $^{1}$Institute of Big Data, College of Computer Science and Artificial Intelligence,\\
  Fudan University, China \\
  $^{2}$Shanghai University of Finance and Economics\\
  \texttt{henghui.ding@gmail.com}
}

\begin{document}

\maketitle

\begin{abstract}
    \vspace{-0.5em}
    Modern video object segmentation (VOS) relies on a two-stage inference paradigm: target initialization and memory-based propagation. However, existing adversarial attacks on VOS often overlook this structure. To bridge this gap, we propose \textbf{\methodname}, an adversarial framework that \textbf{A}ligns with the \textbf{U}niversal two-stage paradigm of \textbf{V}OS models. Specifically, \methodname optimizes a universal adversarial perturbation (UAP) through decoupled training strategies tailored for both initialization and propagation stages. By disrupting the shared memory space, it achieves prompt-agnostic attacks across different modalities, including masks, points, boxes, and language. Extensive experiments show that \methodname establishes state-of-the-art VOS attack performance, such as degrading mask-based SAM3 to only 8.4 $\mathcal{J}\&\mathcal{F}$ on DAVIS17-val (compared to 55.4 for the previous SOTA under the same training setup). It also exhibits superior temporal dynamics with the fastest onset, the most enduring persistence, and the support of any-time mid-video attacks. The \methodname optimization framework is applicable to different VOS models, revealing vulnerabilities in real-world streaming VOS. Code is available at \href{https://github.com/FudanCVL/AUV}{\texttt{https://github.com/FudanCVL/AUV}}.
\end{abstract}


\section{Introduction}

\input{intro}

\section{Related works}
\subsection{Promptable Video Segmentation}
Promptable Video Segmentation focuses on segmenting explicitly specified target(s) within videos, which includes Video Object Segmentation (VOS) \cite{oh2019video, yang2020collaborative, wang2021swiftnet, MOVE, cheng2021rethinking, li2022recurrent, cheng2022xmem, cheng2023tracking, cheng2024putting}, Promptable Concept Segmentation (PCS) \cite{carion2025sam, ding2025mevis, MeViS, GREx, GRES, ding2026multimodal, VLT, ding2021vision, ren2024grounded, li2025sam3, shen2026sam3}, and interactive video segmentation \cite{cheng2021modular, miao2020memory, yang2023track, ravi2024sam, shen2026sam}. Regardless of their differences, these tasks follow a \textbf{consistent two-stage inference paradigm}: prompt-based target initialization and memory-based temporal propagation. The former processes diverse prompts such as masks, points, boxes, and language to encode target identity, while the latter propagates memory-conditioned target information throughout the sequence for consistent video segmentation. This shared structure motivates an attack design tailored to the distinct roles of initialization and propagation.

\subsection{Adversarial Attacks on Segmentation Models}
Adversarial attacks on segmentation models add pixel-level perturbations to input images to disrupt segmentation predictions. Existing image-level methods \cite{hendrik2017universal} either disrupt image features to erode target identity \cite{jiang2024cross, zheng2024black}, or suppress output masks through tailored strategies \cite{zhang2023attack, zhou2024darksam, gu2022segpgd, song2026segtrans, chen2022semantically, rony2023proximal}. Recently, adversarial attacks have been extended to video segmentation models \cite{jiang2023exploring, zhou2025vanish, li2023adversarial, yao2023attention}, either targeting only the initial frame \cite{li2023adversarial, jiang2023exploring} or using a shared optimization objective for all frames within the video sequence \cite{yao2023attention, zhou2025vanish}.

Meanwhile, recent works have also explored ``cross-prompt'' attacks with promising results \cite{zhou2025vanish, liu2024cross, huang2024segment}. However, cross-prompt transferability demonstrated with points or boxes does not necessarily imply generalization to other prompt modalities, particularly dense masks or language, which are processed through distinct encoding pathways. In this work, we aim for a prompt-agnostic attack that remains effective regardless of prompt modalities by corrupting the shared latent memory space.

\section{Methodology}

\input{method}

\section{Experiments}
\label{sec:experiments}

\subsection{Implementation Details}
\label{sec:implementation}
Following prior works \cite{moosavi2017universal, deng2020universal, zhou2024darksam, zhou2025vanish}, we set the perturbation bound to $\epsilon = 10/255$ for all methods. We also provide results with smaller $\epsilon$ in the appendix. Following UAP-SAM2~\cite{zhou2025vanish}, we train all methods on SAM3 using the YTVOS18 training set ~\cite{xu2018youtube} for 10 epochs. For a fair comparison, we sample 8 frames per sequence and use the same learning-rate schedule across all methods. Training is conducted on a single NVIDIA RTX A6000 GPU with 48\,GB of VRAM.

\subsection{Evaluation Metrics} 
\label{sec:metrics}
Following prior works \cite{zhou2025vanish, jiang2023exploring, zhou2024darksam}, we use the standard VOS metrics \cite{perazzi2016benchmark} to evaluate adversarial performance. Furthermore, while these metrics reflect overall attack effectiveness, they fail to characterize the \textit{temporal dynamics} of an attack, such as how quickly the attack takes effect or how long it could persist. To capture these aspects, we further introduce three temporal metrics: \textbf{Latency} ($Lat$), \textbf{Coverage} ($Cov$), and \textbf{Delayed-Coverage} ($Cov_k$).

Let $\mathcal{F}_v$ denote the set of \emph{valid frames} successfully tracked by the benign VOS model. Specifically, a frame is considered valid if its benign score (e.g. $\mathcal{J}\&\mathcal{F}$) exceeds $0.3$, following \cite{zhou2025vanish}. Among these valid frames, we define the set of \emph{collapsed frames} $\mathcal{F}_c$ as those whose score under attack drops by at least $50\%$ relative to the benign score. 

Based on these two sets, we define:  
$\bullet$~\textbf{Latency} ($Lat$), which measures the normalized temporal position of the first collapsed frame in $\mathcal{F}_v$ and thus reflects how quickly the attack takes effect; 
$\bullet$~\textbf{Coverage} ($Cov$), which measures the proportion of collapsed frames in $\mathcal{F}_v$ and thus reflects the temporal persistence of the attack; and 
$\bullet$~\textbf{Delayed-Coverage} ($Cov_k$), which measures the attack coverage $Cov$ for the mid-video attack starting from frame $k$. Formally, these metrics are defined as:

\vspace{-1em}

\begin{equation}
\hspace*{-2.5em}
Lat=
\begin{cases}
\dfrac{\min_{t\in\mathcal{F}_c}\mathrm{rank}_{\mathcal{F}_v}(t)}{|\mathcal{F}_v|}, & \mathcal{F}_c\neq\emptyset,\\[4pt]
1, & \mathcal{F}_c=\emptyset,
\end{cases}
\qquad
Cov=\dfrac{|\mathcal{F}_c|}{|\mathcal{F}_v|},
\qquad
Cov_k=\dfrac{\left|\{\,t\in\mathcal{F}_c\mid t\ge k\,\}\right|}{\left|\{\,t\in\mathcal{F}_v\mid t\ge k\,\}\right|},
\end{equation}

where $\mathrm{rank}_{\mathcal{F}_v}(t)$ denotes the 1-based temporal position
of frame $t$ within $\mathcal{F}_v$, and $k$ denotes the frame of mid-video attack injection. Lower $Lat$ reflects faster onset, while higher $Cov$ and $Cov_k$ indicate better persistence. Low $Cov$ implies potential target recovery during propagation.

\begin{table*}[!t]
\centering
\captionsetup{justification=centering,singlelinecheck=false}
\caption{Attack performance on SAM3~\cite{carion2025sam} across VOS benchmarks and prompt modalities.\\
{\small (All applicable baselines are retrained on SAM3 under the same settings.)}}
\label{tab:full_results_1}
\vspace{-0.5em}
\footnotesize
\renewcommand{\arraystretch}{1}
\resizebox{\textwidth}{!}{
\begin{tabular}{l|ccc|ccc|ccc|ccc}
\toprule
\multirow{2}{*}{Methods} & \multicolumn{3}{c|}{DAVIS17 ($\mathcal{J}\&\mathcal{F}$$\downarrow$)} & \multicolumn{3}{c|}{YTVOS19 ($\mathcal{G}$$\downarrow$)} & \multicolumn{3}{c|}{MOSEv2 ($\mathcal{J}\&\dot{\mathcal{F}}$$\downarrow$)} & \multicolumn{3}{c}{MOSEv1 ($\mathcal{J}\&\mathcal{F}$$\downarrow$)} \\ \cmidrule{2-13}
& Mask & Point & Box & Mask & Point & Box & Mask & Point & Box & Mask & Point & Box \\ \midrule
Clean & 92.2 & 72.1 & 89.1 & 89.7 & 80.0 & 86.8 & 60.3 & 55.8 & 58.9 & 78.4 & 70.1 & 77.0 \\
UAPGD \cite{deng2020universal} & 64.1 & 21.5 & 57.4 & 58.8 & 29.3 & 52.9 & 35.9 & 29.1 & 35.0 & 49.7 & 35.4 & 47.8 \\
VOSPGD \cite{jiang2023exploring} & 61.2 & 19.8 & 54.5 & 49.5 & 21.8 & 42.5 & 35.0 & 28.1 & 33.5 & 44.5 & 30.5 & 42.1 \\
AttackSAM \cite{zhang2023attack} & 68.3 & 30.2 & 63.2 & 63.0 & 30.9 & 53.2 & 42.6 & 36.4 & 38.2 & 52.5 & 41.2 & 47.3 \\
DarkSAM \cite{zhou2024darksam} & 58.5 & 25.0 & 52.5 & 51.7 & 23.1 & 45.6 & 32.5 & 26.9 & 32.0 & 43.0 & 28.0 & 41.3 \\
UAP-SAM2 \cite{zhou2025vanish} & 55.4 & 23.9 & 47.4 & 46.3 & 22.0 & 35.8 & 30.4 & 25.5 & 29.1 & 36.4 & 22.5 & 31.3 \\ \midrule
\rowcolor{defaultColor} \textbf{\methodname (Ours)} & \ \ \textbf{8.4} &  \ \ \textbf{7.1} & \ \ \textbf{7.6} & \textbf{14.0} & \textbf{10.3} & \textbf{12.7} & \textbf{24.6} & \textbf{19.9} & \textbf{23.5} & \textbf{21.1} & \textbf{18.3} & \textbf{20.2} \\ \bottomrule
\end{tabular}
}
\end{table*}

\begin{table*}[t]
\centering
\captionsetup{justification=centering,singlelinecheck=false}
\caption{Temporal dynamics of the attacks on SAM3.}
\vspace{-0.5em}
\label{tab:temporal_analysis_full}
\renewcommand{\arraystretch}{1}
\footnotesize
\setlength{\tabcolsep}{2.2pt} 
\begin{tabular}{l | ccc | ccc | ccc | ccc}
\toprule
\multirow{3}{*}{Method} & \multicolumn{6}{c|}{DAVIS17} & \multicolumn{6}{c}{YTVOS19} \\
\cmidrule(lr){2-7} \cmidrule(lr){8-13}
 & \multicolumn{3}{c|}{Mask} & \multicolumn{3}{c|}{Point} & \multicolumn{3}{c|}{Mask} & \multicolumn{3}{c}{Box} \\
\cmidrule(lr){2-4} \cmidrule(lr){5-7} \cmidrule(lr){8-10} \cmidrule(lr){11-13}
 & $Lat$$\downarrow$ & $Cov$$\uparrow$ & $Cov_{20}$$\uparrow$ & $Lat$$\downarrow$ & $Cov$$\uparrow$ & $Cov_{20}$$\uparrow$ & $Lat$$\downarrow$ & $Cov$$\uparrow$ & $Cov_{20}$$\uparrow$ & $Lat$$\downarrow$ & $Cov$$\uparrow$ & $Cov_{20}$$\uparrow$ \\
\midrule
SegPGD \cite{gu2022segpgd} & 53.7 & 36.5 & 45.7 & 10.0 & 86.4 & 87.5 & 53.9 & 38.6 & 46.4 & 46.6 & 46.6 & 55.3 \\
DarkSAM \cite{zhou2024darksam} & 58.3 & 33.5 & 41.6 & 11.0 & 84.6 & 85.0 & 56.1 & 36.6 & 43.8 & 46.6 & 44.8 & 52.6 \\
UAP-SAM2 \cite{zhou2025vanish} & 52.5 & 35.9 & 43.6 & 10.4 & 85.1 & 85.2 & 50.9 & 41.1 & 47.4 & 36.1 & 56.6 & 63.5 \\
\midrule
\rowcolor{defaultColor} \textbf{\methodname (Ours)} & \ \ \textbf{5.5} & \textbf{89.2} & \textbf{88.1} & \ \ \textbf{2.2} & \textbf{90.3} & \textbf{88.6} & \ \ \textbf{8.6} & \textbf{85.1} & \textbf{84.6} & \ \ \textbf{7.0} & \textbf{86.8} & \textbf{85.9} \\
\bottomrule
\end{tabular}
\end{table*}

\subsection{Main Results}

\begin{figure*}[!t]
    \centering
    \includegraphics[width=1.0\textwidth]{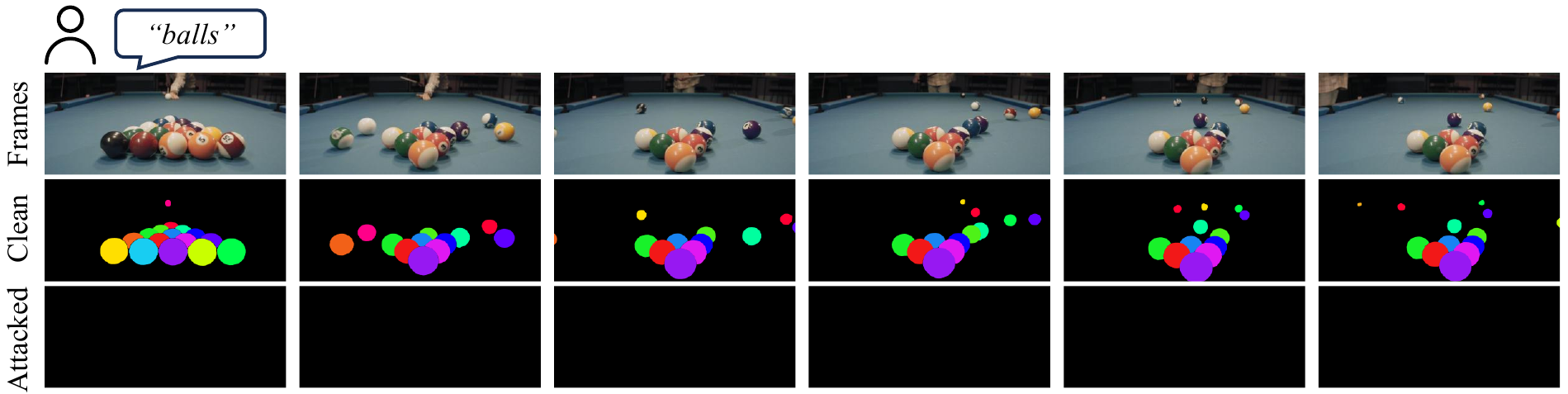}
    \vspace{-1.5em}
    \caption{\methodname simultaneously suppresses multiple targets specified by a single language prompt.}
    \label{fig:figure4}
    \vspace{-0.5em}
\end{figure*}

\begin{table*}[!t]
\centering
\caption{Attack performance on SAM3 across language-based SA-Co/VEval benchmarks \cite{carion2025sam}.}
\vspace{-0.5em}
\renewcommand{\arraystretch}{1}
\label{tab:full_results_2}
\resizebox{\textwidth}{!}{
\begin{tabular}{l|cccc|cccc|cccc}
\toprule
\multirow{2}{*}{Methods} & \multicolumn{4}{c|}{SA-V} & \multicolumn{4}{c|}{YT-Temporal-1B} & \multicolumn{4}{c}{SmartGlasses} \\ \cmidrule{2-13}
& cgF1 & pHOTA & pDetA & pAssA & cgF1 & pHOTA & pDetA & pAssA & cgF1 & pHOTA & pDetA & pAssA \\ \midrule
Clean & 29.3 & 60.7 & 44.7 & 83.2 & 50.2 & 70.5 & 60.5 & 82.7 &  33.5 & 60.2 &  46.2 & 79.3 \\
UAD \cite{lu2024unsegment} &  \ \ 2.4 & 24.9 & \ \ 8.5 & 74.0 & \ \ 7.8 & 31.7 & 15.0 & 67.5 & 11.1 & 36.6 & 21.0 & 64.4 \\
DarkSAM \cite{zhou2024darksam} & \ \ 4.0 & 30.5 & 12.6 & 75.0 & 10.7 & 38.8 & 22.7 & 67.0 & 12.4 & 38.8 & 23.0 & 65.9 \\
UAP-SAM2 \cite{zhou2025vanish} & \ \ 1.7 & 23.9 & \ \ 7.8 & 74.6 & \ \ 3.5 & 26.2 & 11.1 & 62.4 & \ \ 9.1 & 36.3 & 21.4 & 62.4 \\ \midrule
\rowcolor{defaultColor} \textbf{\methodname (Ours)} & \ \ \textbf{0.1} & \ \ \textbf{8.5} & \ \ \textbf{1.2} & \textbf{64.3} & \ \ \textbf{0.3} & \ \ \textbf{9.1} & \ \ \textbf{1.7} & \textbf{49.9} & \ \ \textbf{1.9} & \textbf{18.4} & \ \ \textbf{6.4} & \textbf{53.4} \\ \bottomrule
\end{tabular}
}
\vspace{-0.5em}
\end{table*}

\textbf{Overall Attack Performance.} 
We evaluate \methodname on the validation sets of seven benchmarks across four prompt modalities (masks, points, boxes, and language) on SAM3 \cite{carion2025sam}. As shown in Tables \ref{tab:full_results_1} and \ref{tab:temporal_analysis_full}, \methodname consistently outperforms prior attacks across benchmarks and prompt types, such as reducing mask-based $\mathcal{J}\&\mathcal{F}$ on DAVIS17 to only 8.4, while achieving the fastest attack onset ($Lat$) and the highest coverage ($Cov$). MOSE is excluded from temporal analysis because per-frame ground-truth masks are unavailable. \methodname also generalizes to language prompts, achieving 0.1 cgF1 and 8.5 pHOTA on the SA-V split of SA-Co/VEval (Table~\ref{tab:full_results_2}) and enabling one-to-many attacks that simultaneously suppress multiple objects specified by a single language prompt (Figure~\ref{fig:figure4}). See appendix for more quantitative comparisons and visualizations.

\textbf{Applicability of \methodname to other VOS models.} 
Table \ref{tab:vos_models_comparison} demonstrates that the \methodname framework can also be applied to other VOS models under the same optimization procedure, including SAM2 \cite{ravi2024sam}, Cutie \cite{cheng2024putting} and XMem \cite{cheng2022xmem}. For example, it reduces the mask-based $\mathcal{J}\&\mathcal{F}$ of SAM2.1-B+ on DAVIS17 from 89.0 to 12.1, and the $\mathcal{G}$ of Cutie-base on YTVOS19 from 87.0 to 22.4, demonstrating that the proposed two-stage attack framework is applicable across different VOS models.

\begin{table*}[!t]
\centering
\scriptsize
\renewcommand{\arraystretch}{1.1}
\setlength{\tabcolsep}{2.5pt}
\setlength{\aboverulesep}{0pt}
\setlength{\belowrulesep}{0pt}
\caption{Attack performance on different VOS models and benchmarks using mask prompts.
}
\vspace{-0.5em}
\label{tab:vos_models_comparison}
\begin{tabular}{l|c|c|ccc|ccccc|ccc|ccc}
\toprule
\multirow{2}{*}{VOS models} & \multirow{2}{*}{Venue} & \multirow{2}{*}{State} & \multicolumn{3}{c|}{DAVIS17} & \multicolumn{5}{c|}{YTVOS19} & \multicolumn{3}{c|}{MOSEv2} & \multicolumn{3}{c}{MOSEv1} \\ \cmidrule{4-17}
 & & & $\mathcal{J}\&\mathcal{F}$ & $\mathcal{J}$ & $\mathcal{F}$ & $\mathcal{G}$ & $\mathcal{J}_s$ & $\mathcal{F}_s$ & $\mathcal{J}_u$ & $\mathcal{F}_u$ & $\mathcal{J}\&\dot{\mathcal{F}}$ & $\mathcal{J}$ & $\dot{\mathcal{F}}$ & $\mathcal{J}\&\mathcal{F}$ & $\mathcal{J}$ & $\mathcal{F}$ \\ \midrule
\multirow{2}{*}{XMem \cite{cheng2022xmem}} & \multirow{2}{*}{\pub{ECCV'22}} & Clean & 86.2 & 82.9 & 89.5 & 85.5 & 84.3 & 88.6 & 80.3 & 88.6 & 36.3 & 34.7 & 37.9 & 57.6 & 53.3 & 62.0 \\
 & & Adv & 24.0 & 22.4 & 25.6 & 26.3 & 26.2 & 28.4 & 24.1 & 26.5 & 28.3 & 27.5 & 29.1 & 25.7 & 24.8 & 26.6 \\ \hline
\multirow{2}{*}{Cutie-base \cite{cheng2024putting}} & \multirow{2}{*}{\pub{CVPR'24}} & Clean & 87.9 & 84.6 & 91.1 & 87.0 & 86.0 & 90.5 & 82.0 & 89.6 & 42.8 & 41.1 & 44.4 & 69.9 & 65.9 & 74.1 \\
 & & Adv & 18.3 & 17.5 & 19.1 & 22.4 & 23.6 & 21.0 & 22.2 & 22.8 & 21.1 & 20.6 & 21.6 & 22.2 & 21.3 & 23.1 \\ \hline
\multirow{2}{*}{SAM2.1-T \cite{ravi2024sam}} & \multirow{2}{*}{\pub{ICLR'25}} & Clean & 88.8 & 85.4 & 92.2 & 87.4 & 85.5 & 90.0 & 83.0 & 91.3 & 45.9 & 44.1 & 47.8 & 71.6 & 67.4 & 75.9 \\
 & & Adv & 17.8 & 17.2 & 18.4 & 25.0 & 25.4 & 23.8 & 24.6 & 26.2 & 20.6 & 19.8 & 21.4 & 24.5 & 23.4 & 25.6 \\ \hline
\multirow{2}{*}{SAM2.1-B+ \cite{ravi2024sam}} & \multirow{2}{*}{\pub{ICLR'25}} & Clean & 89.0 & 86.0 & 92.0 & 88.0 & 86.0 & 90.4 & 83.6 & 91.8 & 46.0 & 44.2 & 47.8 & 73.6 & 69.6 & 77.6 \\
 & & Adv & 12.1 & 12.4 & 11.8 & 19.2 & 22.8 & 17.6 & 18.5 & 17.9 & 26.5 & 26.4 & 26.7 & 23.2 & 22.5 & 23.8  \\ \hline
\multirow{2}{*}{SAM2.1Long-B+ \cite{ding2025sam2long}} & \multirow{2}{*}{\pub{ICCV'25}} & Clean & 89.4 & 86.4 & 92.5 & 88.1 & 86.0 & 90.5 & 84.0 & 91.8 & 47.4 & 45.5 & 49.2 & 74.7 & 70.6 & 78.8 \\
 & & Adv & 15.9 & 16.0 & 15.7 & 23.0 & 26.5 & 21.2 & 22.0 & 22.3 & 25.2 & 25.0 & 25.4 & 23.4 & 22.4 & 24.3 \\ \bottomrule
\end{tabular}
\vspace{-1.5em}
\end{table*}

\subsection{Two-Stage Joint Optimization}
\savegeometry{-1em}
We validate our two-stage optimization on SAM3 by comparing it with two variants: training only on the initial frame (Stage I / Init. Only) or training only on subsequent frames (Stage II / Sub. Only).

\textbf{Attack performance across inference settings.} Table \ref{tab:two-stage} compares their performance under three different \textit{inference-time} attack settings: attacking only the initial frame ({Init.}); attacking only the subsequent frames ({Sub.}); or attacking the entire sequence ({All}). As shown, training on either stage does not generalize well to the other. For instance, the UAP trained only on subsequent frames has limited effectiveness when applied to the initial frame, with SAM3 retaining a 70.1 $\mathcal{J}$\&$\mathcal{F}$ on DAVIS17. This reveals the intrinsic gap between the initialization and propagation stages of VOS models and supports our main motivation. With two-stage joint optimization, \methodname achieves strong attack performance across all three settings, outperforming both single-stage-trained variants.

\textbf{Temporal dynamics.} Table \ref{tab:two-stage-ttc-tic} compares the temporal dynamics of three training configurations. The $Cov_{20}$ results show that training only on the initial frame generalizes poorly to mid-video attacks, which is increasingly important with the rise of online VOS methods~\cite{bekuzarov2023xmem++,cheng2018fast, wu2023scalable}. In contrast, our two-stage training achieves both faster onset ($Lat$) and higher coverage ($Cov$), along with strong mid-video attack performance ($Cov_{20}$), demonstrating superior temporal dynamics. 

\begin{figure*}[!t]
    \centering
    \includegraphics[width=0.4\textwidth]{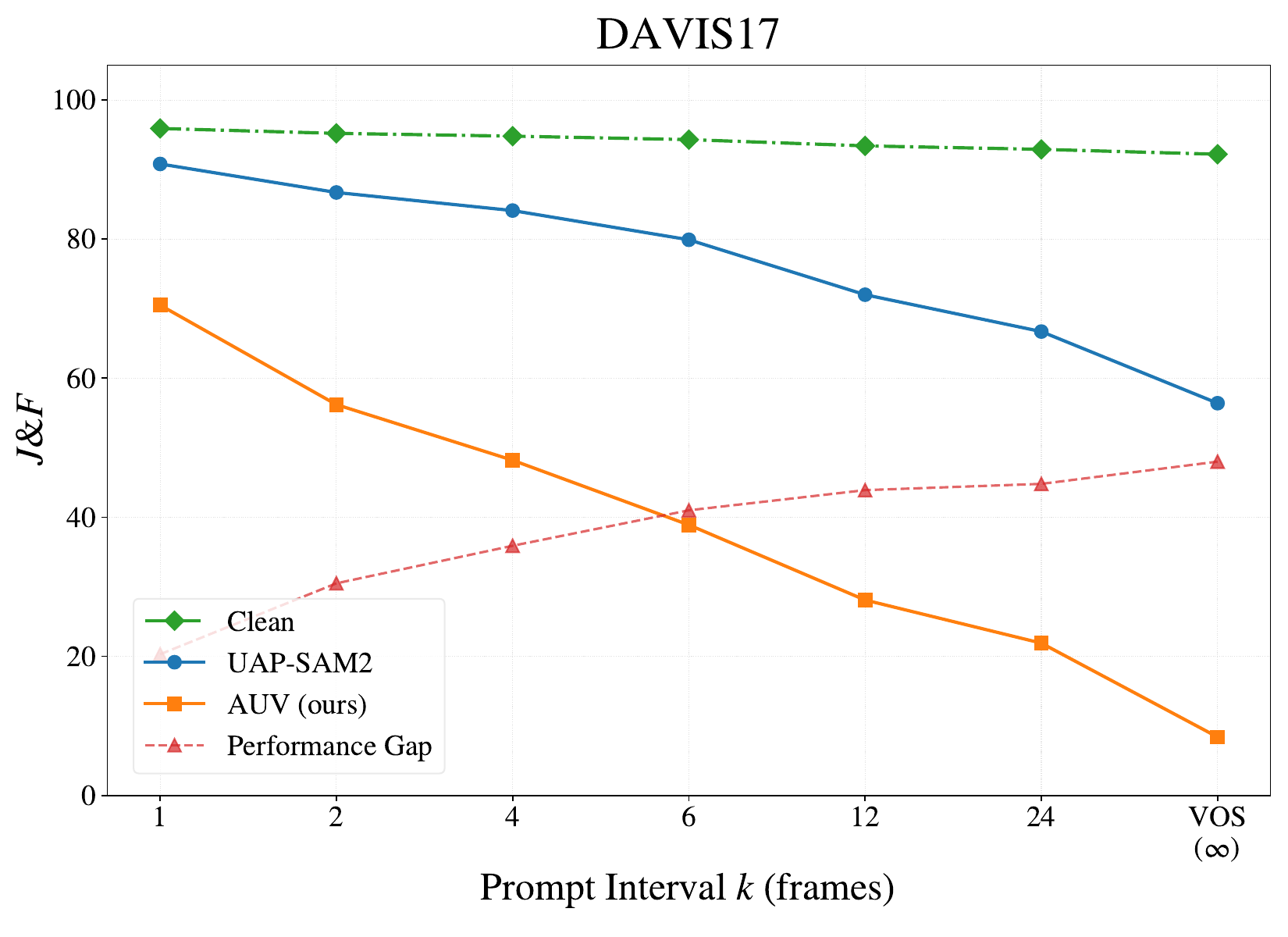}
    \includegraphics[width=0.4\textwidth]{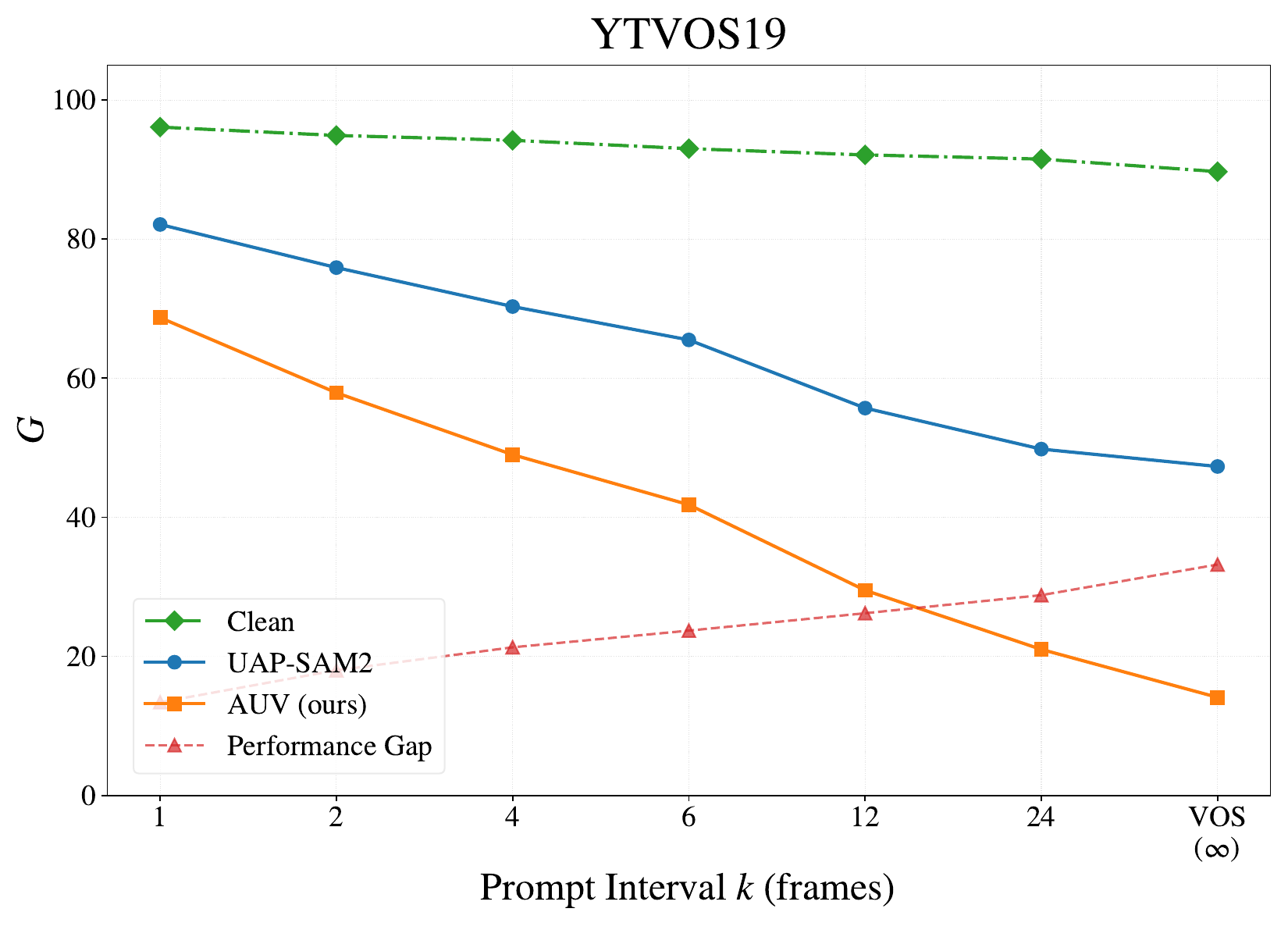}
    \vspace{-2mm}
    \caption{Comparison of attack performance under different prompting intervals $k$, where a larger $k$ shifts the model's reliance from prompt-based initialization to memory-based propagation.}
    \label{fig:figure5}
    \vspace{-1em}
\end{figure*}

\textbf{From prompt-based initialization to memory-driven propagation.} 
We compare \methodname with the previous SOTA, UAP-SAM2 \cite{zhou2025vanish}, under periodic online correction, providing a ground-truth mask every $(k+1)$ frames for $k \in \{1,2,4,6,12,24,\infty \}$. Here, $k=1$ means the model is corrected every other frame, and $k=\infty$ corresponds to the standard VOS, with a mask provided only at initialization. Larger $k$ increases reliance on memory-driven propagation and reduces human intervention. Figure~\ref{fig:figure5} shows that \methodname consistently outperforms UAP-SAM2 across all $k$ values, from frequent ground-truth corrections to pure memory-driven propagation, with an even larger advantage in the latter, highlighting its sustained effectiveness throughout the VOS propagation stage.

\begin{table}[!t]
    \centering
    \footnotesize
    \begin{minipage}[t]{0.52\textwidth}
        \centering
        \caption{Ablation on training stages and inference settings. Init.: Initial frame; Sub.: Subsequent frames.}
        \label{tab:two-stage}
        \vspace{0.4em}
        \renewcommand{\arraystretch}{1}
        \setlength{\tabcolsep}{2pt} 
        \begin{tabular}{l|ccc|ccc|ccc}
            \toprule
            \multirow{2}{*}{\diagbox[width=1.5cm, height=0.9cm]{Train}{Infer.}} & \multicolumn{3}{c|}{DAVIS17~\textsubscript{\tiny $\mathcal{J}$\&$\mathcal{F}$}} & \multicolumn{3}{c|}{YTVOS19~\textsubscript{\ $\mathcal{G}$}} & \multicolumn{3}{c}{MOSEv1~\textsubscript{\tiny $\mathcal{J}$\&$\mathcal{F}$}} \\ \cmidrule(lr){2-4} \cmidrule(lr){5-7} \cmidrule(lr){8-10}
            & Init. & Sub. & All & Init. & Sub. & All & Init. & Sub. & All \\ \midrule
            Init. Only & 47.2 & 23.9 & 26.2 & 46.0 & 24.4 & 29.4 & 43.2 & 20.9 & 21.4 \\ 
            Sub. Only & 70.1 & 13.6 & 28.9 & 57.4 & 16.4 & 29.8 & 52.8 & \textbf{18.4} & 23.0 \\
            \rowcolor{defaultColor} \textbf{Init. + Sub.} & \textbf{23.3} & \ \ \textbf{9.6} & \ \ \textbf{8.4} & \textbf{24.1} & \textbf{15.3} & \textbf{14.0} & \textbf{29.5} & 20.3 & \textbf{21.1} \\ \bottomrule
        \end{tabular}
    \end{minipage}
    \hfill 
    \begin{minipage}[t]{0.46\textwidth}
        \centering
        \caption{Ablation on temporal dynamics: $Lat$ (Latency), $Cov$ (Coverage), and $Cov_k$.}
        \label{tab:two-stage-ttc-tic}
        \vspace{0.4em}
        \renewcommand{\arraystretch}{1}
        \setlength{\tabcolsep}{1pt}
        \begin{tabular}{@{}l|ccc|ccc@{}}
            \toprule
            \multirow{2}{*}{\diagbox[width=1.5cm, height=0.9cm]{Train}{\hspace{0.4cm}Infer.}}
            & \multicolumn{3}{c|}{DAVIS17} & 
            \multicolumn{3}{c}{YTVOS19} \\ \cmidrule(lr){2-4} \cmidrule(lr){5-7} 
            & $Lat$$\downarrow$ & $Cov$$\uparrow$ & $Cov_{20}$$\uparrow$ & $Lat$$\downarrow$ & $Cov$$\uparrow$ & $Cov_{20}$$\uparrow$ \\ \midrule
            Init. Only & 21.3 & 69.9 & 68.5 & 26.5 & 64.4 & 69.5 \\ 
            Sub. Only & 21.4 & 64.9 & 87.8 & 29.1 & 62.2 & \textbf{84.8} \\
            \rowcolor{defaultColor} \textbf{Init. + Sub.} & \ \ \textbf{5.5} & \textbf{89.2} & \textbf{88.1} & \ \  \textbf{8.6} & \textbf{85.1} & 84.6 \\ \bottomrule
        \end{tabular}
    \end{minipage}
    \vspace{-0.5em}
\end{table}

\begin{table}[!t]
\centering
\caption{Prompt-agnostic capability of \methodname by training on one prompt type and evaluating on other unseen prompt types. $^\dagger$: Method's default setting.}
\label{tab:cross_prompt}
\vspace{0.2em}
\footnotesize
\setlength{\tabcolsep}{5.4pt}
\renewcommand{\arraystretch}{0.9}
\begin{tabular}{@{}llccccccccc@{}}
\toprule
\multirow{2}{*}{Method} & \multirow{2}{*}{Training} & \multicolumn{3}{c}{DAVIS17 ($\mathcal{J}\&\mathcal{F}$$\downarrow$)} & \multicolumn{3}{c}{YTVOS19 ($\mathcal{G}$$\downarrow$)} & SA-V & YT-1B & SG \\ \cmidrule(lr){3-5} \cmidrule(lr){6-8} \cmidrule(l){9-11} 
                                 &                   & Mask & Point & Box & Mask & Point & Box & \multicolumn{3}{c}{Language (pHOTA$\downarrow$)} \\ \midrule
\multirow{2}{*}{UAP-SAM2 \cite{zhou2025vanish}} & Scanning Points$^\dagger$ & 55.4 & 23.9 & 47.4 & 46.3 & 22.0 & 35.8 & 23.9 & 26.2 & 36.3 \\
& Box  &   62.1   &   21.8   &  52.5  &    50.2  &  21.8  &  38.8  & 27.7  &   34.0   &  40.4  \\ 
\midrule
\multirow{3}{*}{\methodname (Ours)} & Mask           & \ \  \textbf{8.4} & \ \  7.1 & \ \  7.6 & \textbf{14.0} & 10.3 & 12.7 & \ \ \textbf{8.5} & \ \  \textbf{9.1} & \textbf{18.4} \\
                                 & Point             & 17.7     & \ \  \textbf{5.7} & 11.2     & 19.1 \note{20.3} & \ \ \textbf{9.0} \note{8.1} & 13.0 & 12.2 & 13.8 & 21.5 \note{22.8} \\
                                 & Box               & 15.2     & \ \  7.8 & \ \  \textbf{6.5} & 17.5 & 11.0 & \textbf{11.1} & 11.7     & 12.4 & 20.2 \\ \bottomrule
\end{tabular}
\vspace{-0.5em}
\end{table}

\subsection{Prompt-Agnostic Attack}
\label{sec:prompt_agnostic}
\vspace{-0.5em}
\textbf{Overall prompt-agnostic generalization.} We investigate the prompt-agnostic capability of \methodname on SAM3 by \textbf{training each variant exclusively on one prompt modality and evaluating it zero-shot on other unseen modalities}. As shown in Table \ref{tab:cross_prompt}, \methodname remains highly effective regardless of the prompt type used to specify the target during training. For instance, the mask-trained UAP reduces point-based $\mathcal{J}\&\mathcal{F}$ to 7.1 on DAVIS17, while the box-trained UAP drops mask-based $\mathcal{G}$ to 17.5 on YTVOS19. This prompt-agnosticism extends to language prompts, with mask-trained UAP reaching 9.1 pHOTA on YT-1B of SA-Co/VEval. In contrast, existing method \cite{zhou2025vanish} suffers from a significant modality gap, showing degradation on dense masks due to the sparse-branch-specific optimization.

\begin{figure*}[t]
    \centering
    \begin{subfigure}[t]{0.30\textwidth}
        \centering
        \includegraphics[width=\linewidth]{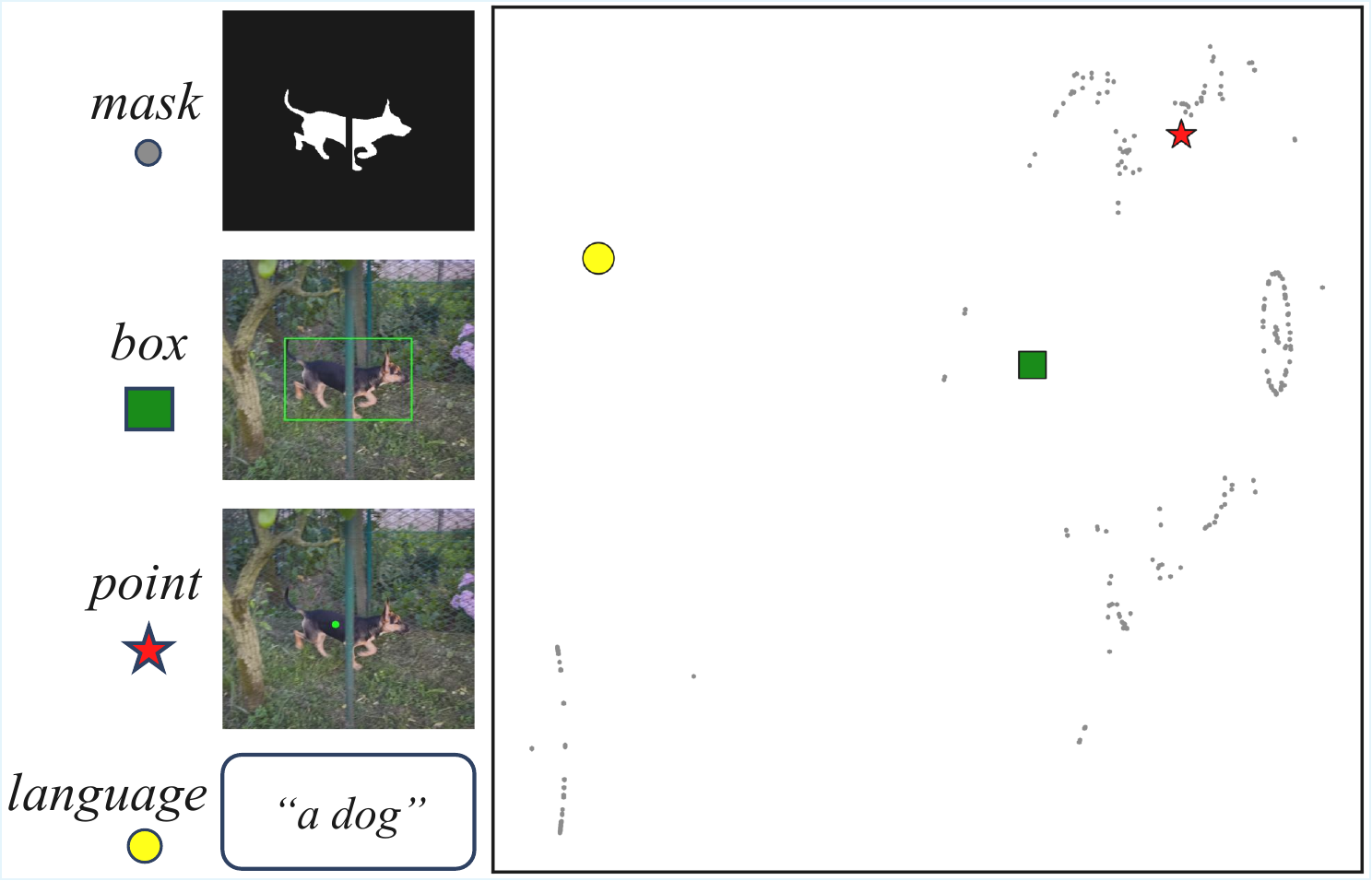}
        \vspace{-5mm}
        \caption{\scriptsize Prompt embeddings.}
        \label{fig:bypass_a}
    \end{subfigure}
    \hfill
    \begin{subfigure}[t]{0.26\textwidth}
        \centering
        \includegraphics[width=\linewidth, height=3.4cm]{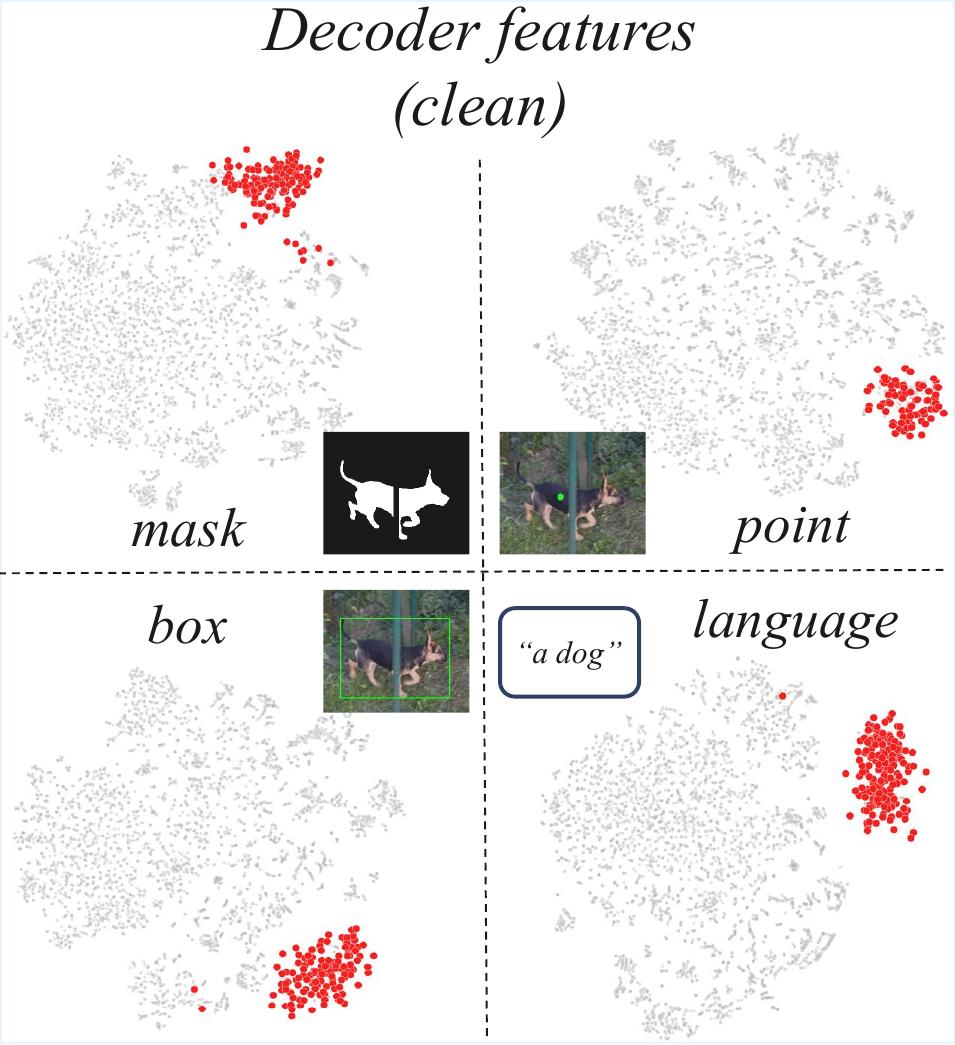}
        \vspace{-5mm}
        \caption{\scriptsize Decoder features.}
        \label{fig:bypass_b}
    \end{subfigure}
    \hfill
    \begin{subfigure}[t]{0.37\textwidth}
        \centering
        \includegraphics[width=\linewidth]{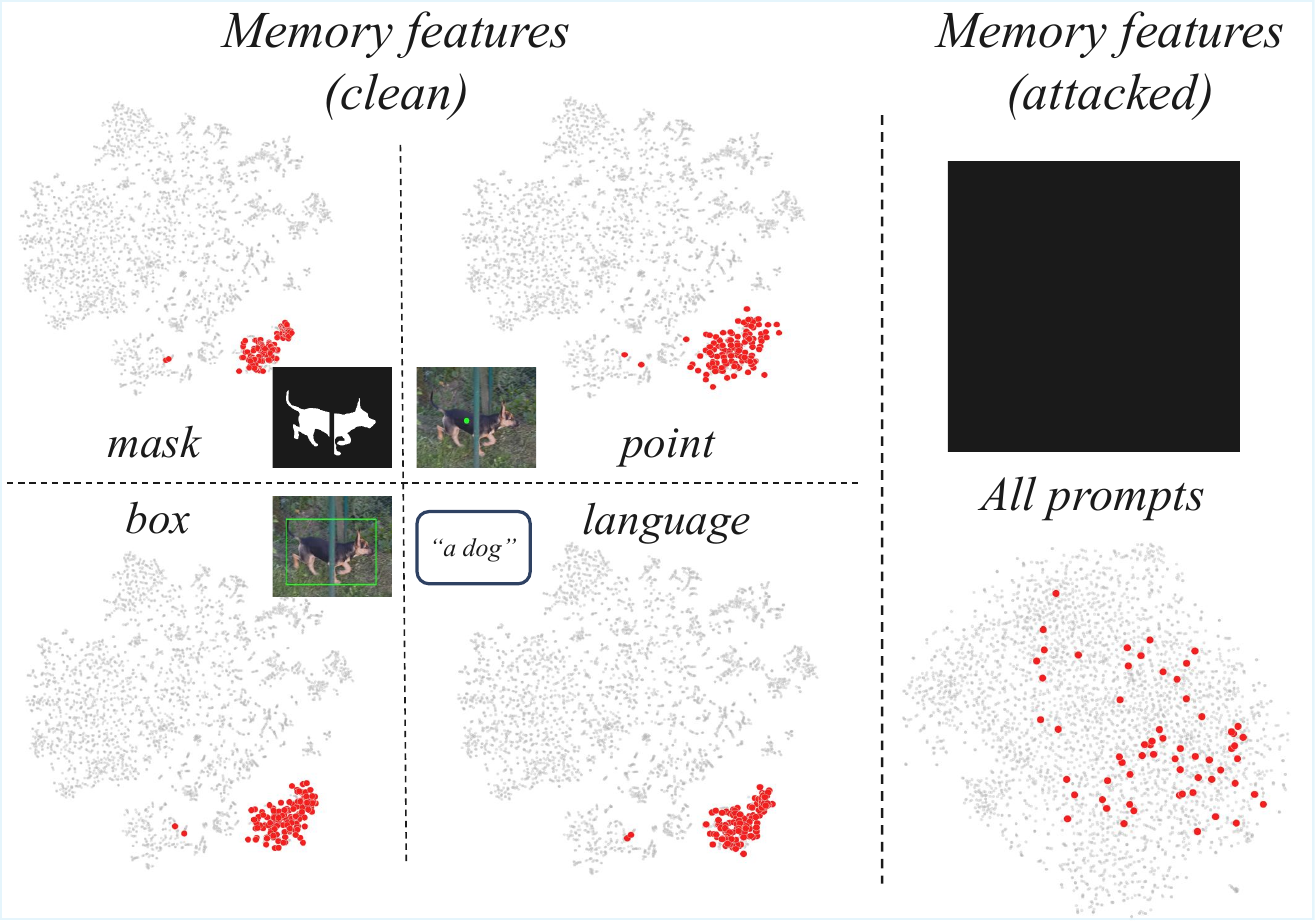}
        \vspace{-5mm}
        \caption{\scriptsize Memory features.}
        \label{fig:bypass_c}
    \end{subfigure}
    \caption{t-SNE visualizations of latent features: (a) prompt embeddings; (b) final-block decoder features; and (c) memory features before (left) and after (right) the attack. \textcolor{red}{Foreground} and \textcolor{gray}{background} features in ground-truth are marked in \textcolor{red}{red} and \textcolor{gray}{gray}, respectively.}
    \label{fig:latent_investigation}
    \vspace{-1em}
\end{figure*}

\textbf{Investigating the root cause of prompt-agnosticism.} Figure~\ref{fig:bypass_a} visualizes the t-SNE~\cite{van2008visualizing} map of different prompt embeddings, where distinct prompt modalities form separated clusters. Accordingly, the decoder features used for mask prediction remain prompt-conditioned and exhibit different distributions, as shown in Figure~\ref{fig:bypass_b}. Consequently, prior attacks targeting predicted masks can become tied to specific prompt-encoding branches, and therefore, struggle to generalize to other prompts processed by distinct branches (e.g., from sparse points to dense masks). In contrast, the distributions of memory features are highly aligned across prompt modalities (Figure~\ref{fig:bypass_c}, left), suggesting that prompt-specific variations are largely filtered out in this shared space while common target cues are retained. By attacking this shared latent space (Figure~\ref{fig:bypass_c}, right), \methodname entangles foreground and background features, thus collapsing predictions across all prompt modalities.

To further reveal the mechanism behind this effect, we quantitatively analyze all videos across multiple VOS benchmarks. For each video, we compute pairwise cross-prompt cosine similarities separately for decoder and memory features using mask, point, and box prompts. As reported in Table~\ref{tab:cross_prompt_similarity}, decoder-feature similarities can be as low as 0.6 to 0.7, revealing clear modality differences, particularly between dense (mask) and sparse (point \& box) prompts. In contrast, memory-feature similarities consistently range from 0.98 to 1.00, confirming that the \textbf{memory space remains highly consistent regardless of whatever prompts}. This reveals the basis of \methodname's prompt-agnosticism, which allows a UAP trained with one prompt modality to transfer zero-shot to others.

\begin{table}[t]
\caption{Pairwise cross-prompt cosine similarities of decoder and memory features, averaged over all videos of each benchmark.}
\label{tab:cross_prompt_similarity}
\vspace{0.2em}
\centering\small
\setlength{\tabcolsep}{7pt}
\renewcommand{\arraystretch}{1.15}
\begin{tabular}{@{}l*{6}{c}@{}}
\toprule
& \multicolumn{3}{c}{Decoder Features}
& \multicolumn{3}{c}{Memory Features $E_1$} \\
\cmidrule(lr){2-4}\cmidrule(l){5-7}
Benchmark & Mask--Point & Mask--Box & Point--Box
        & Mask--Point & Mask--Box & Point--Box \\
\midrule
DAVIS17 & 0.780 & 0.682 & 0.860 & 0.983 & 0.997 & 0.985 \\
YTVOS19 & 0.784 & 0.679 & 0.852 & 0.984 & 0.993 & 0.988 \\
MOSEv1  & 0.772 & 0.626 & 0.852 & 0.991 & 0.998 & 0.992 \\
MOSEv2  & 0.761 & 0.611 & 0.848 & 0.993 & 0.999 & 0.994 \\
\bottomrule
\end{tabular}
\end{table}

\begin{figure*}[!t]
    \centering
    \begin{subfigure}[b]{0.245\textwidth}
        \centering
        \includegraphics[width=\linewidth]{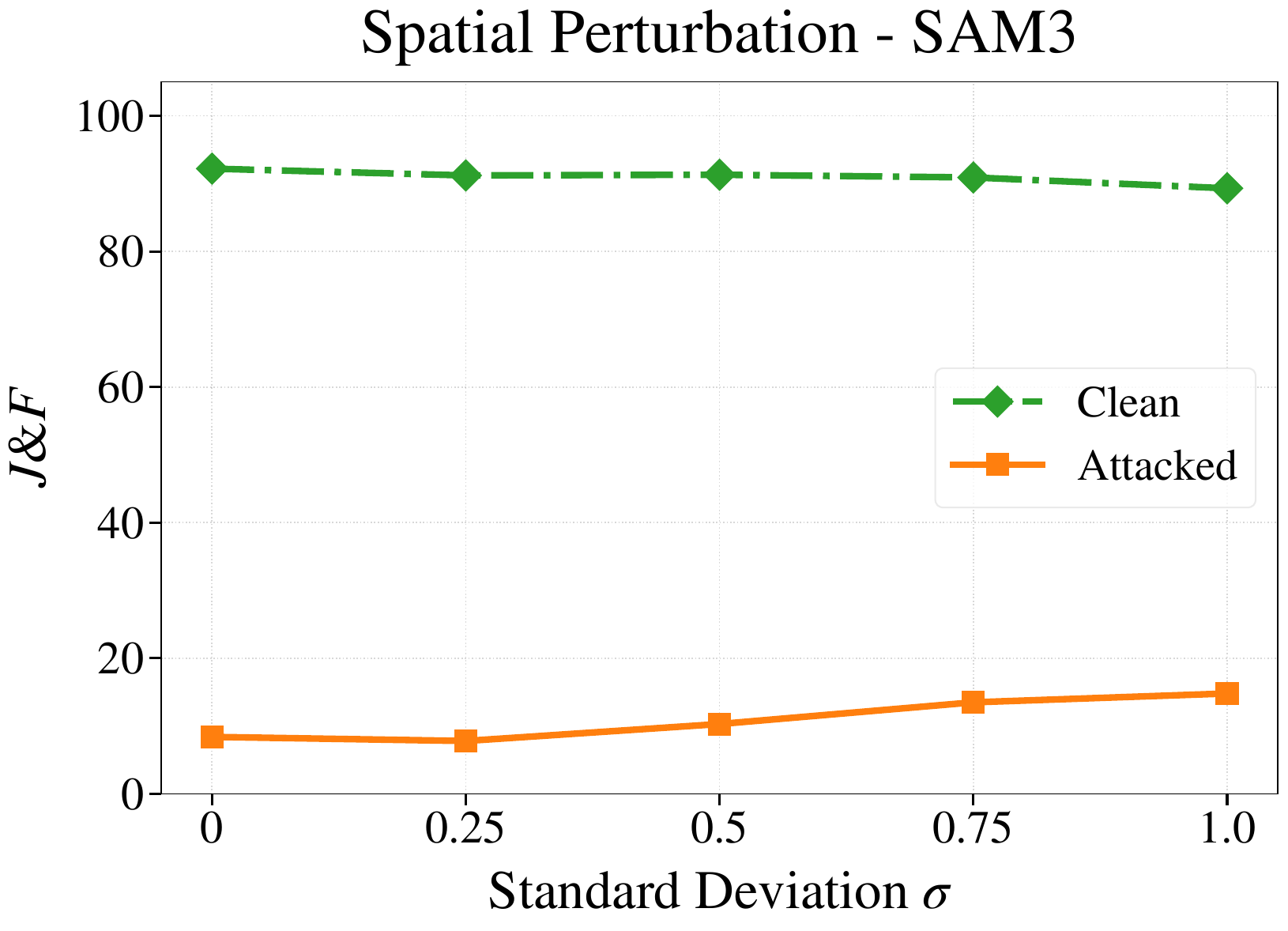}
        \vspace{-6mm}
        \caption*{\scriptsize (a)}
    \end{subfigure}
    \begin{subfigure}[b]{0.245\textwidth}
        \centering
        \includegraphics[width=\linewidth]{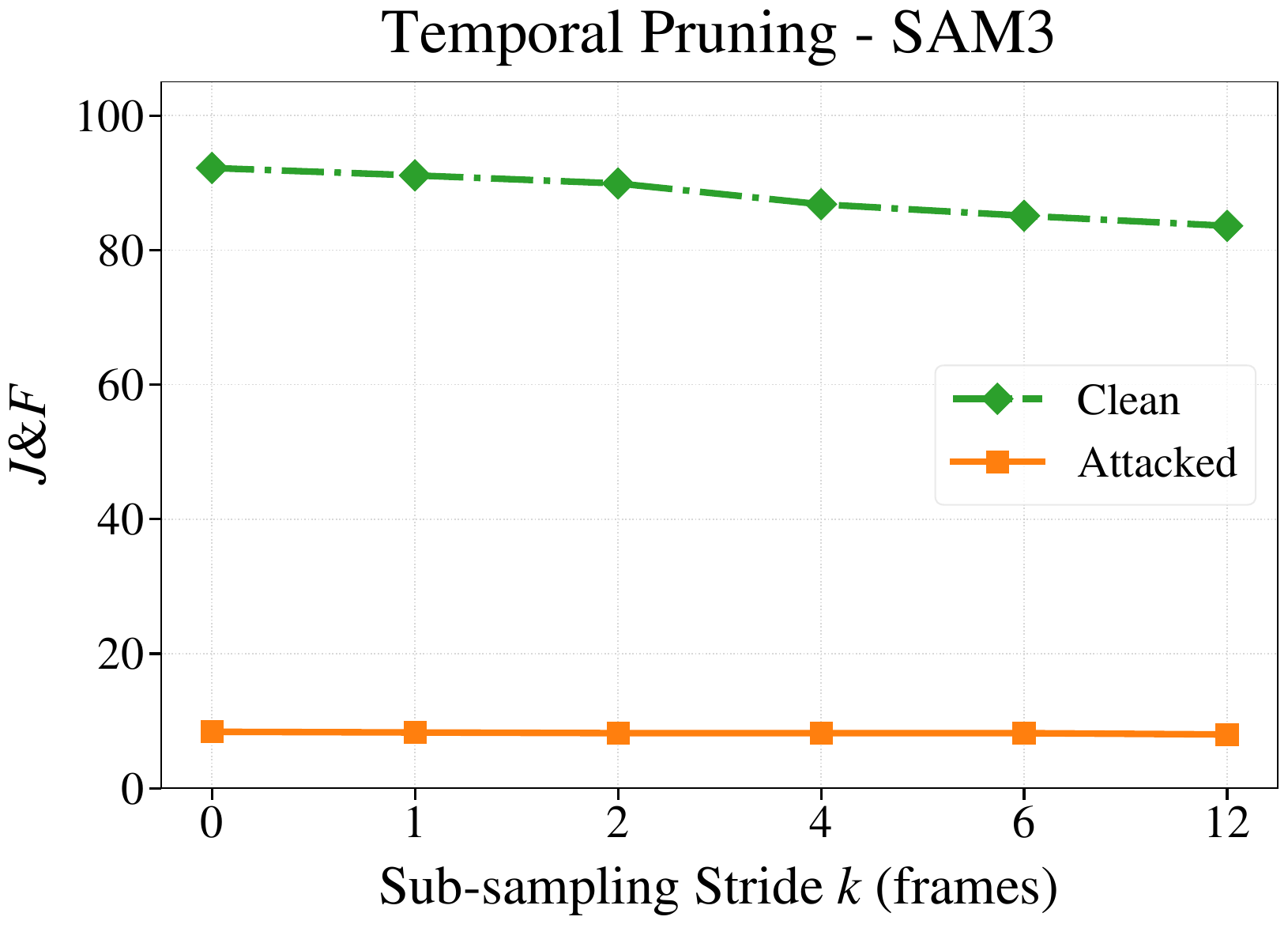}
        \vspace{-6mm}
        \caption*{\scriptsize (b)}
    \end{subfigure}
    \begin{subfigure}[b]{0.245\textwidth}
        \centering
        \includegraphics[width=\linewidth]{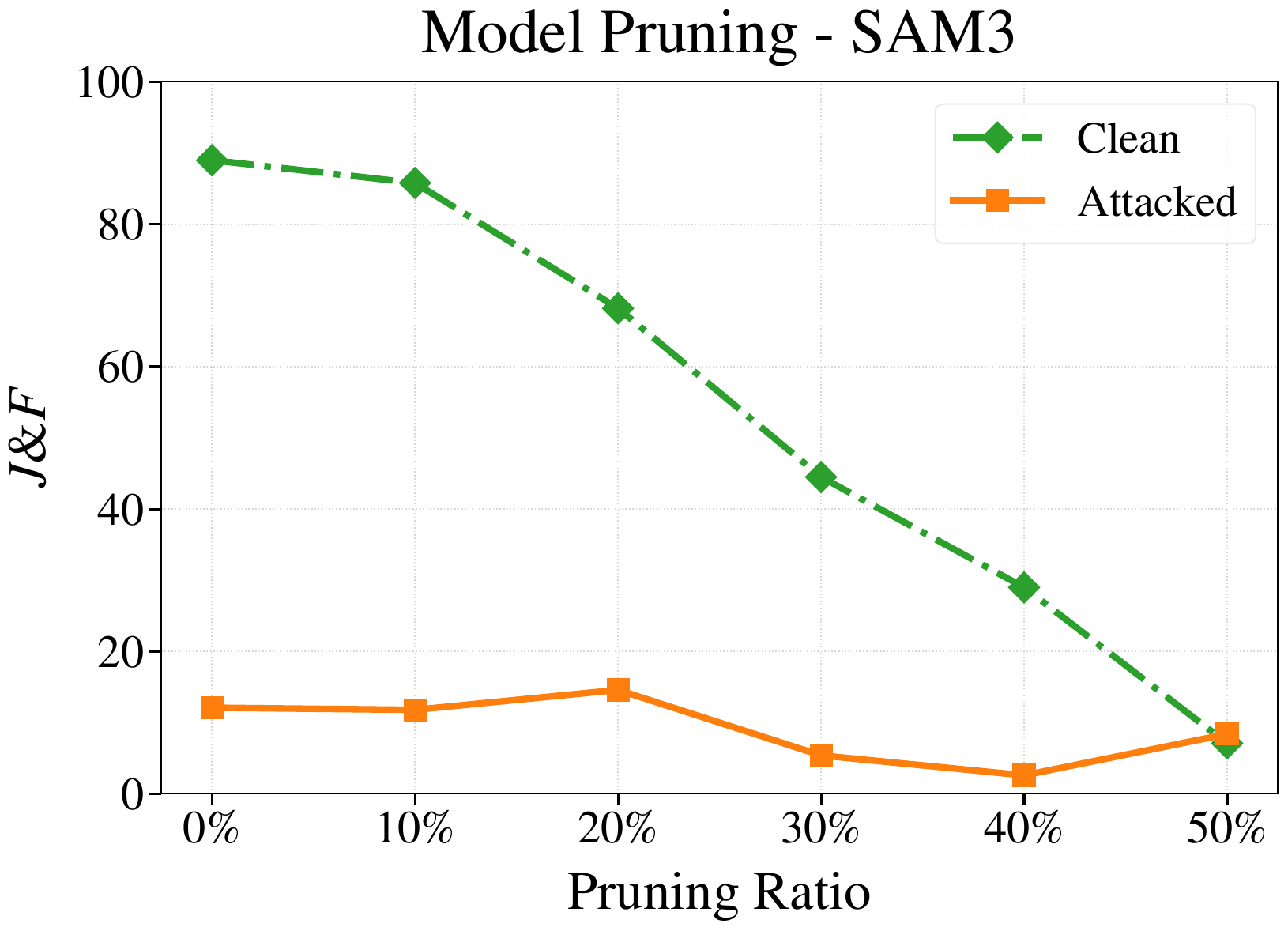}
        \vspace{-6mm}
        \caption*{\scriptsize (c)}
    \end{subfigure}
    \begin{subfigure}[b]{0.245\textwidth}
        \centering
        \includegraphics[width=\linewidth]{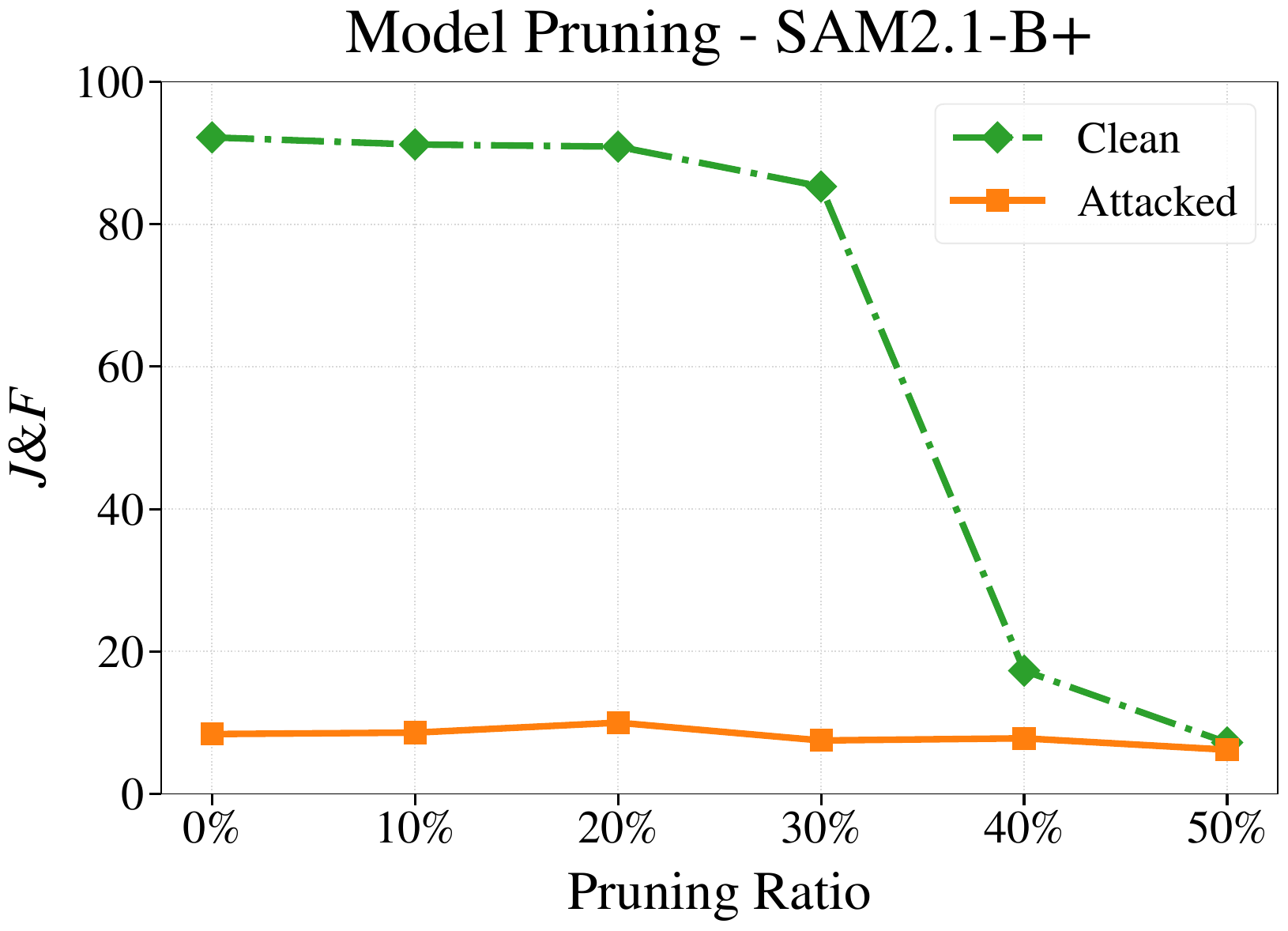}
        \vspace{-6mm}
        \caption*{\scriptsize (d)}
    \end{subfigure}
    \vspace{-1.5em}
    \caption{Defense study on spatial perturbation (a), temporal pruning (b), and model pruning (c,d).}
    \label{fig:defense}
    \vspace{-1em}
\end{figure*}

\subsection{Defense Study}
\label{sec:defense}
\vspace{-0.5em}
We evaluate the robustness of \methodname against potential defenses. All experiments are conducted on the DAVIS17 validation set for a controlled comparison.

\textbf{Spatial perturbation.} We distort images using Gaussian blur with standard deviation $\sigma \in \{0.25, 0.5, 0.75, 1.0\}$. This smooths local image details, with larger $\sigma$ producing stronger blur. As shown in Figure~\ref{fig:defense}a, \methodname remains robust across different spatial perturbation levels.

\textbf{Temporal pruning.} We sub-sample every $k$-th frame within each video sequence, discarding the remaining frames to reduce frame rate and disrupt temporal continuity of videos. As shown in Figure \ref{fig:defense}b, \methodname's attack efficacy remains stable, demonstrating robustness to temporal discontinuity.

\textbf{Model pruning.} Following \cite{zhu2017prune, zhou2025vanish}, we apply magnitude-based unstructured pruning. At each pruning ratio, the smallest-magnitude weights in each targeted layer are set to zero. As in Figure~\ref{fig:defense}c,d, \methodname maintains attack effectiveness on both SAM3 and SAM2.1-B+ across different sparsity levels.

\textbf{Periodic re-prompting.} As shown in Figure \ref{fig:figure5}, we provide a ground-truth mask every $(k+1)$ frames to simulate user corrections for target recovery. \methodname remains effective across all tested intervals.

\textbf{Memory-level defenses.} We further evaluate random dropping of non-conditioning memory entries and more advanced memory-management strategies. \methodname remains effective under both settings; detailed results are provided in the appendix.

\section{Limitations and Broader Impact}
\vspace{-0.5em}
\methodname assumes white-box access to the target VOS model during offline UAP optimization, which is consistent with prior works that require gradient-based source-model access \cite{hendrik2017universal, zhang2023attack, zhou2025vanish}, and therefore does not currently support black-box optimization. Once optimized, the fixed UAP requires no further model access, prompt, or per-video optimization. Meanwhile, \methodname is primarily designed for memory-based VOS and may be less effective on earlier architectures without explicit memory mechanisms. The attack could potentially be misused to disrupt deployed VOS systems; we therefore emphasize its intended use only for robustness evaluation and scientific research. It can also serve as a tool for assessing VOS robustness to adversarial perturbations and guiding the development of more resilient models.

\section{Conclusion}
\vspace{-0.5em}
We propose \methodname, a universal adversarial attack framework for VOS models. \methodname aligns its optimization with the two-stage VOS inference paradigm by decoupling UAP training into target initialization and memory propagation stages with tailored supervisions. By corrupting the shared memory space, it achieves strong prompt-agnostic attacks across masks, points, boxes, and language. Extensive experiments demonstrate \methodname's state-of-the-art attack performance and the framework's broad applicability across VOS models. It delivers the fastest attack onset and superior persistence, while supporting flexible mid-video attacks, providing a strong baseline for future research.

\section*{Acknowledgements}
This work was supported by the National Natural Science Foundation of China (NSFC) under Grant No. 62472104 and the Science and Technology Commission of Shanghai Municipality under Grant No.~25511103600, and Shanghai Pujiang Program 2025PJA201. Henghui Ding was supported by Xiaomi Young Scholars Program.
{
\small

\bibliographystyle{plain}
\bibliography{reference}

\newpage
\input{supple.tex}

\end{document}

%% file: intro.tex
\textbf{Modern video object segmentation (VOS) \cite{oh2019video, cheng2024putting, ravi2024sam, carion2025sam} systems follow a two-stage inference paradigm.} In the first stage, the target is specified through a prompt, such as a mask, point, box, or language. In the second stage, the model propagates the target across frames using memory, with little or no further prompting. This two-stage design is foundational to recent VOS models such as XMem \cite{cheng2022xmem}, Cutie \cite{cheng2024putting}, SAM2 \cite{ravi2024sam}, SAM3 \cite{carion2025sam}, and is central to their robust performance.

\begin{figure*}[!h]
    \centering
    \includegraphics[width=1.0\textwidth, height=5.7cm]{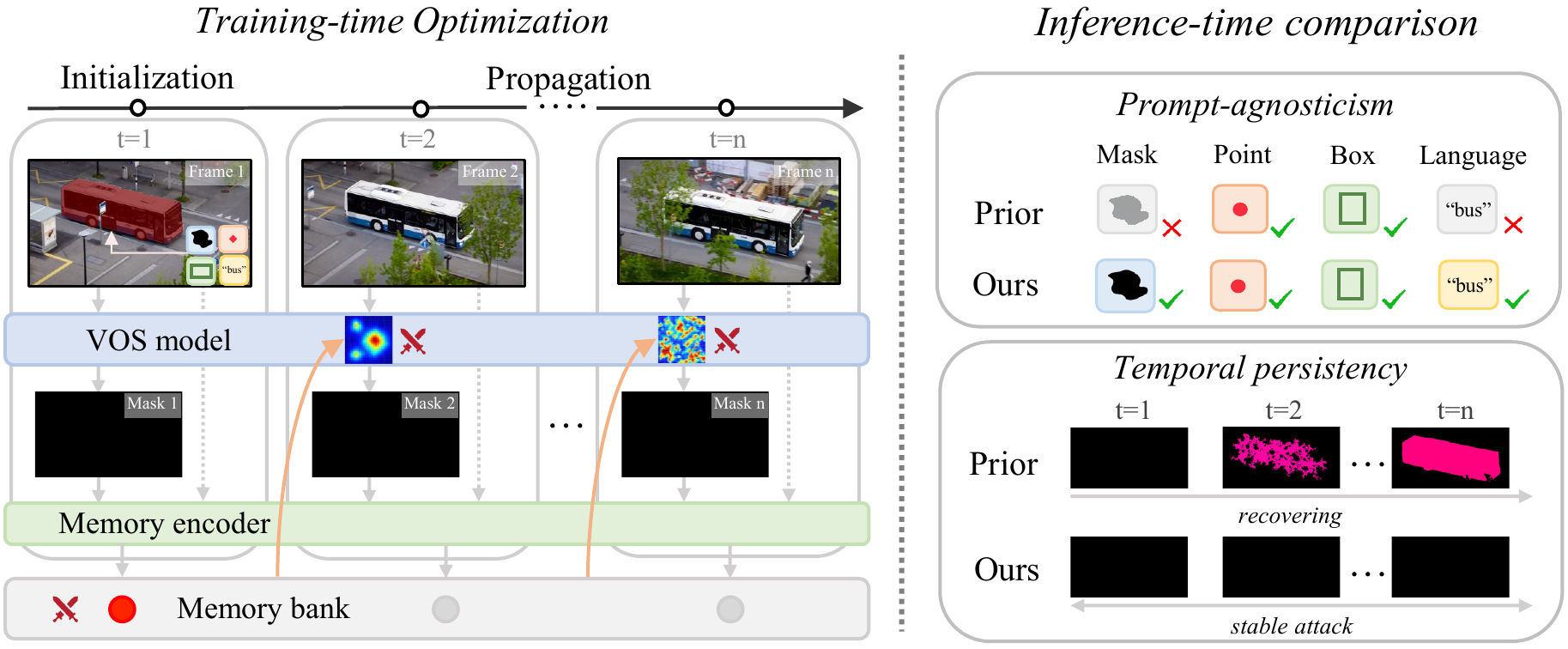}
    \caption{\methodname aligns attack optimization with the two-stage VOS inference pipeline (initialization and propagation), yielding more persistent and prompt-agnostic attacks than prior methods.}
    \label{fig:figure1}
\end{figure*}

Meanwhile, despite the rapid development of adversarial attacks, including those targeting VOS models \cite{jiang2023exploring, zhou2025vanish, li2023adversarial, yao2023attention}, \textbf{existing methods suffer from a fundamental training-inference misalignment}: they typically optimize the attacks under a single-stage formulation. Specifically, they either focus on the prompt-conditioned initial frame while neglecting memory-based temporal propagation~\cite{li2023adversarial, jiang2023exploring}, or apply the same optimization objective for all frames during training~\cite{zhou2025vanish, yao2023attention}. This mismatch between \textbf{a uniform attack optimization strategy and the two-stage VOS inference pipeline} often leads to two critical limitations: \textbf{(1)} the attacks do not effectively corrupt latent memory representations stored in the memory bank, potentially allowing the model to recover the target during propagation; \textbf{(2)} they often target the predicted mask, a prompt-biased by-product in VOS. Consequently, attacks optimized with sparse points or boxes often struggle to transfer to other prompts encoded by different branches, such as dense masks or language.

Motivated by these observations, we propose \textbf{\methodname}, a stage-\textbf{A}ligned \textbf{U}niversal \textbf{V}OS attack framework that optimizes a universal adversarial perturbation (UAP) to disrupt both the initialization and propagation stages of the target VOS model. As illustrated in Figure \ref{fig:figure1} (left), \methodname decouples the attack optimization into two complementary stages that align with the VOS inference pipeline. \textbf{During prompt-based initialization}, it optimizes the UAP to corrupt the target's memory feature anchor before it is permanently stored in the memory bank, thereby erasing the target identity at its source. \textbf{During memory-based propagation}, it optimizes the \textbf{same} UAP to flatten the spatial saliency of memory-attended feature maps, pushing their activations toward spatial uniformity and thereby severing temporal target association across frames. 

Our paradigm-aligned design not only delivers stronger attacks, with faster onset and more sustained temporal persistence, but also supports a broader range of prompt modalities and more flexible attack timing, as illustrated in Figure \ref{fig:figure1} (right). First, by targeting the shared latent memory space, \methodname avoids prompt-specific optimization and transfers zero-shot across different prompt modalities, including masks, points, boxes, and language. Second, by targeting the propagation stage, \methodname supports any-time mid-video attacks, where the perturbation can be injected from any intermediate frame and remains effective throughout the subsequent sequence.

Extensive experiments validate our design. Under the same training setup and perturbation budget, \methodname consistently outperforms prior attacks across VOS benchmarks and prompt modalities on SAM3, with the performance gap even more pronounced during pure memory-based propagation. Moreover, the optimization framework of \methodname applies naturally to different VOS models, suggesting it exploits shared properties of modern VOS systems, rather than a model- or prompt-specific weakness. The main contributions are summarized as follows:

\begin{itemize}

    \item \textbf{Structural Alignment:} We identify a fundamental mismatch between existing uniform attack optimizations and the two-stage VOS inference pipeline. To address this, we propose \methodname, a stage-aligned UAP-based attack framework that simultaneously corrupts memory anchor during initialization and flattens memory-attended features during propagation.
        
    \item \textbf{Prompt-Agnostic Attack:} Our attack targets the shared latent memory space rather than prompt-conditioned output masks, enabling zero-shot generalization across different prompt modalities, including mask, point, box, and language.
    
    \item \textbf{Metrics \& Empirical Insights:} We introduce new metrics to evaluate the onset speed and persistence of video attacks. Extensive experiments show \methodname's new SOTA attack performance, while further analyses provide insights into its underlying mechanisms.
    
\end{itemize}

%% file: method.tex
\label{sec:method}

\subsection{Preliminaries and Problem Formulation}
\label{sec:prelim}

A video sequence is denoted as $\mathcal{V} = \{I_1, I_2, \dots, I_T\}$, where $I_i \in \mathbb{R}^{H \times W \times 3}$. Given a target-specifying prompt $P$ (mask, point, box, or language) applied to the initial frame, a VOS model $\Phi$ predicts a binary mask sequence $\{M_i\}_{i=1}^T$ with $M_i \in \{0,1\}^{H \times W}$. To attack $\Phi$, we optimize a universal adversarial perturbation (UAP) $\delta \in \mathbb{R}^{H \times W \times 3}$ that can be applied consistently to every frame and is strictly bounded by an $\ell_\infty$ budget $\epsilon$ to keep it invisible:
\begin{equation}
\|\delta\|_\infty \leq \epsilon.
\end{equation}
The adversarial video is $\tilde{\mathcal{V}} = \{I_i + \delta\}_{i=1}^T$, and the attack objective is to maximize the deviation between the adversarial prediction and the clean prediction:
\begin{equation}
\max_{\delta} \sum_{i=1}^T \mathcal{D}(\tilde{M}_i, M_i) \quad \text{s.t.} \quad \|\delta\|_\infty \leq \epsilon,
\label{eq:objective}
\end{equation}
where $\tilde{M}_i$ is the adversarial mask predicted at frame $i$ and $\mathcal{D}(\cdot,\cdot)$ quantifies spatial divergence.

\paragraph{Two-stage VOS inference pipeline.} Modern VOS models implement $\Phi$ through two distinct stages: an \emph{initialization stage} $\Phi_{\text{init}}$ that produces an initial memory feature $E_1$, which is permanently stored in the memory bank $\mathcal{B}$ as an anchor; and a \emph{propagation stage} $\Phi_{\text{prop}}$ that, for each subsequent frame, performs memory readout from $\mathcal{B}$ for temporal target association.

\begin{figure*}[!t]
    \centering
    \includegraphics[width=1.0\textwidth]{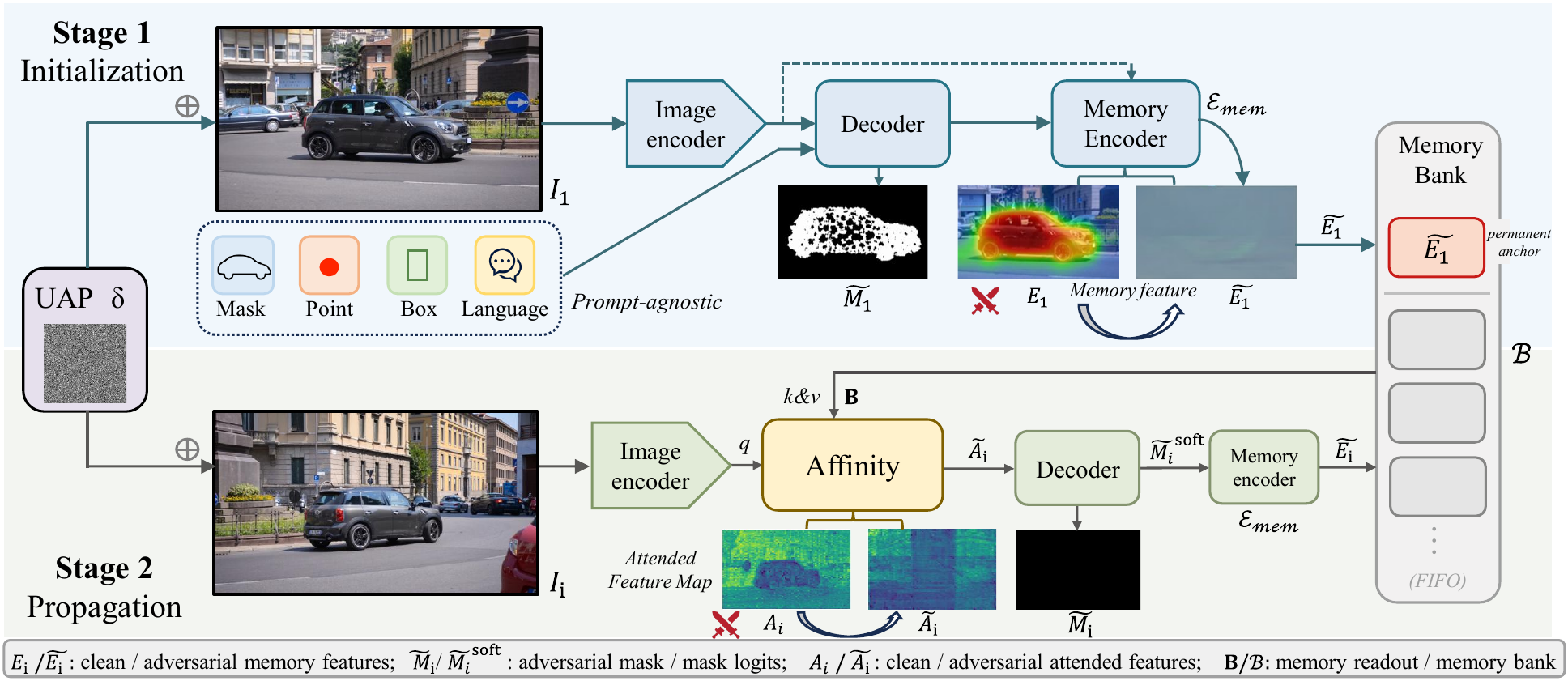}
    \caption{\textbf{Overview of \methodname.} Our UAP-based \methodname is optimized to corrupt the initial memory feature ($\tilde{E}_1$), attacking target identity at its source, and to flatten the spatial saliency of memory-attended feature maps ($\tilde{A}_i$), collapsing target association during propagation. It targets the shared latent memory space for prompt-agnostic attacks and supports attack optimization on different VOS models.}
    \label{fig:figure2}
\end{figure*}

\subsection{Paradigm Misalignment and Method Overview}
\label{sec:overview}
Existing VOS attacks optimize $\delta$ using a uniform or single-stage objective, typically maximizing $\sum_i \mathcal{D}(\tilde{M}_i, M_i)$ either only on the initial frame or on all frames with the same supervision strategy. This ignores the distinct roles of $\Phi_{\text{init}}$ and $\Phi_{\text{prop}}$ in VOS inference. \textbf{We refer to this mismatch between uniform attack optimization and two-stage VOS inference as a \emph{training-inference paradigm misalignment}}, which could insufficiently disrupt both the initial memory anchor that establishes target identity and the memory readout that propagates it throughout the sequence.
\vspace{0.5em}

\paragraph{Design principle.} The overview of our method is shown in Figure~\ref{fig:figure2}. We propose \textbf{\methodname}, whose optimization aligns with the two-stage VOS inference pipeline by decoupling the UAP training into two distinct stages that directly mirror $\Phi_{\text{init}}$ and $\Phi_{\text{prop}}$:

\begin{itemize}
    \item \textbf{Stage I - Initialization-aligned corruption.} We corrupt the initial memory feature $\tilde{E}_1$ \emph{before} it is permanently stored in $\mathcal{B}$ as an anchor, erasing the target identity at its source. The target can be specified at initialization using a prompt $P$ of any modality.

    \item \textbf{Stage II - Propagation-aligned flattening.} We flatten the spatial saliency of memory-attended feature maps $\tilde{A}_i$ in subsequent frames, disrupting the temporal chain of target propagation. The prompt encoder is disabled, consistent with VOS inference where no prompts are provided after initialization.
    The \textbf{same} UAP $\delta$ is shared across both stages.
\end{itemize}

\subsection{Stage I: Corrupting the Permanent Memory Anchor}
\label{sec:stage1}

\begin{figure*}[h!]
  \begin{minipage}[c]{0.60\textwidth}
    \paragraph{Why initialization must be attacked differently.} In $\Phi_{\text{init}}$, the prompt $P$ first produces an initial mask $M_1$, after which a memory encoder $\mathcal{E}_{\text{mem}}$ fuses the image feature $F_1$ with the underlying soft values to obtain a target-specific memory feature $E_1 \in \mathbb{R}^{h \times w \times c}$. Modern VOS models store this $E_1$ in the memory bank as a \textbf{ \emph{permanent anchor}} for the entire sequence. Meanwhile, prior methods typically optimize $\delta$ to corrupt the \emph{output} mask $M_1$ \cite{jiang2023exploring, zhou2025vanish, li2023adversarial, yao2023attention}, an intermediate product in the VOS pipeline. However, an empty binarized output mask $\tilde{M}_1$ does not imply that its underlying soft values $\tilde{M}_1^{\mathrm{soft}}$ used for memory encoding contain no target information. As shown in Fig.~\ref{fig:figure3}, these residual spatial cues keep the target visible in $\tilde{E}_1$ despite the empty output mask, which allows potential recovery during subsequent propagation.
  \end{minipage}
  \hfill
  \begin{minipage}[c]{0.38\textwidth}
    \centering
    \includegraphics[width=\textwidth, height=3.6cm]{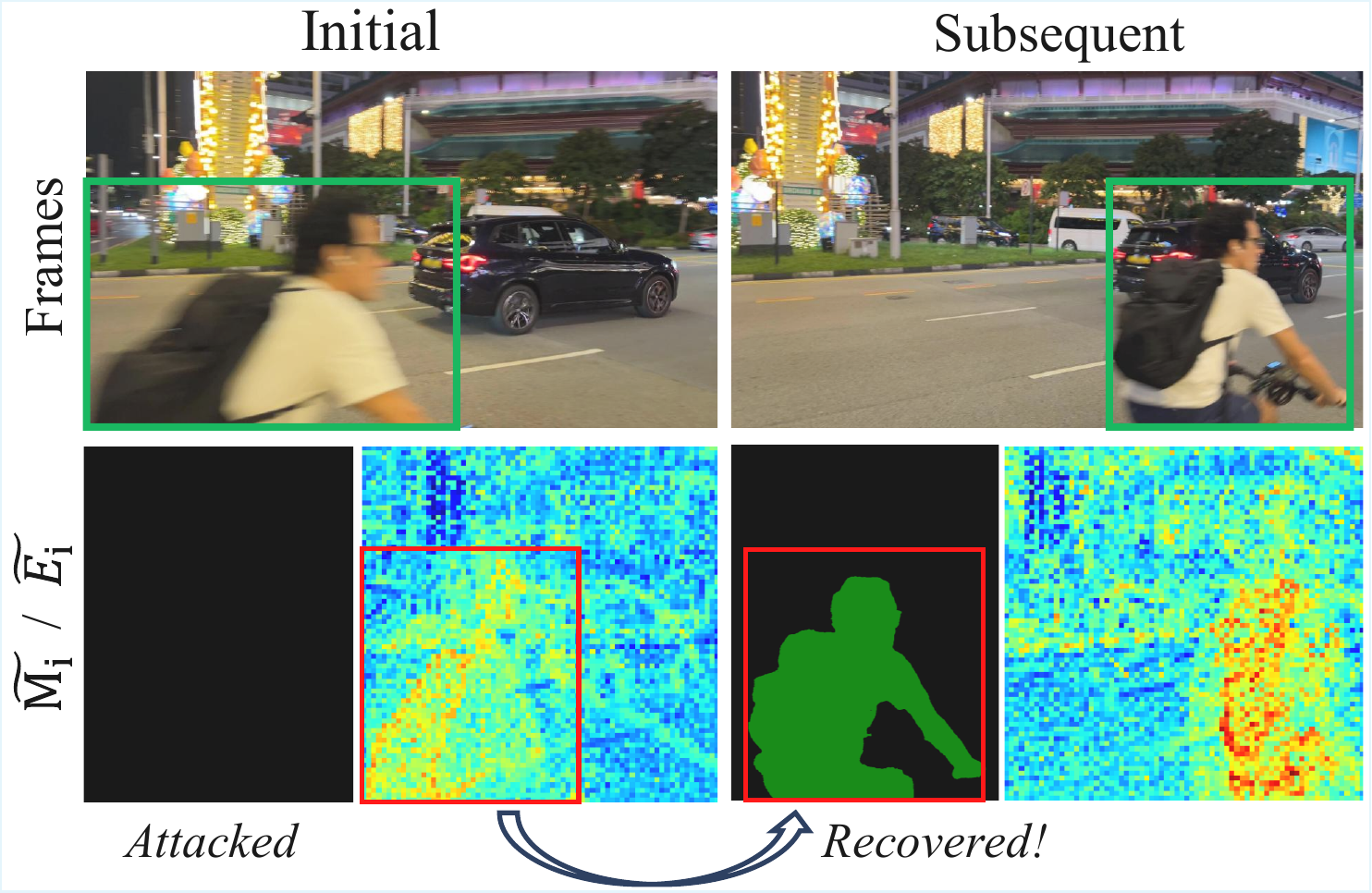}
    \caption{Previous video attacks \cite{jiang2023exploring, li2023adversarial, yao2023attention, zhou2025vanish} suppress mask $\tilde{M}_1$, but $\tilde{E}_1$ remains subtly visible, enabling potential target recovery in later frames.}
    \label{fig:figure3}
  \end{minipage}
\end{figure*}

\paragraph{Direct memory-feature corruption.} We therefore optimize the UAP $\delta$ to directly corrupt the memory feature $E_1$. Specifically, we define three representations: the adversarial feature $\tilde{E}_1 = \mathcal{E}_{\text{mem}}(I_1+\delta, \tilde{M}_1^{\mathrm{soft}})$, the clean feature $E_1 = \mathcal{E}_{\text{mem}}(I_1+\delta, M_1^{\mathrm{soft}})$, and a background-only reference $E_1^\emptyset = \mathcal{E}_{\text{mem}}(I_1+\delta, \mathbf{0})$, where $\mathbf{0}$ denotes an all-zero mask prior with no spatial target cues. The initialization-aligned loss pulls the adversarial feature toward the background state while pushing it away from the clean state:
\begin{equation}
\mathcal{L}_{\text{init}}
= \frac{\|\tilde{E}_1 - E_1^\emptyset\|_2^2}{hwc}
+ \exp\!\left(-\frac{10\|\tilde{E}_1 - E_1\|_2^2}{hwc}\right),
\label{eq:linit}
\end{equation}
where the exponential term pushes the adversarial feature away from the clean feature.

Unlike most prior video attacks that disrupt prompt-conditioned masks ~\cite{jiang2023exploring, li2023adversarial, yao2023attention, zhou2025vanish}, $\mathcal{L}_{\text{init}}$ is formulated in the shared memory feature space to corrupt the permanent memory anchor $\tilde{E}_1$ that preserves the target identity needed for subsequent propagation.

\subsection{Stage II: Flattening Memory-Attended Features}
\label{sec:stage2}

\paragraph{Why propagation must be attacked differently.} For each subsequent frame $i \in \{2,\dots,T\}$, $\Phi_{\text{prop}}$ performs temporal association by cross-attending the image feature $F_i$ to the memory readout $\mathbf{B} \in \mathbb{R}^{N \times h \times w \times c}$, producing a memory-attended feature map $A_i \in \mathbb{R}^{h \times w \times c}$:
\begin{equation}
A_i = \text{softmax}(F_i \mathbf{B}^\top)\mathbf{B} + F_i,
\label{eq:attention}
\end{equation}
where $N$ denotes the number of frames currently stored in the memory bank $\mathcal{B}$. In a benign VOS model, $A_i$ exhibits strong spatial saliency that localizes the target \cite{oh2019video, dosovitskiy2020image}. This saliency is the \textbf{carrier of temporal propagation}: the decoder uses $A_i$ to predict $M_i$, whose soft values are then fused with image features to form a new memory feature stored in $\mathcal{B}$ for subsequent frames.

\paragraph{Propagation erosion via variance minimization.} From this point, we optimize the \textbf{same} $\delta$ to flatten $\tilde{A}_i$ into a spatially uniform feature map, stripping away all target-localized activation peaks. Let $\tilde{\mu}_i \in \mathbb{R}^{1 \times 1 \times c}$ be the spatial mean of $\tilde{A}_i$:
$\tilde{\mu}_i = \frac{1}{hw} \sum_{x=1}^{h}\sum_{y=1}^{w} \tilde{A}_i(x,y,:)$, 
the propagation-aligned loss penalizes spatial variance as follows:
\begin{equation}
\mathcal{L}_{\text{prop}}^{(i)} = \frac{1}{hwc} \sum_{x=1}^{h}\sum_{y=1}^{w} \big\|\tilde{A}_i(x,y,:) - \tilde{\mu}_i\big\|_2^2.
\label{eq:lprop}
\end{equation}
Flattening the saliency of $\tilde{A}_i$ prevents the decoder from recovering the target. The resulting collapsed state is then encoded into $\mathcal{B}$, triggering cascading failure throughout the remaining sequence.

\subsection{Joint Two-Stage Optimization}
\label{sec:optimization}

\paragraph{Staged curriculum.} Directly optimizing the combined loss from scratch is unstable: if the memory anchor $E_1$ is not yet corrupted, Stage II will have to disrupt propagation that still carries a strong clean target signal. We therefore adopt a two-phase curriculum that follows the causal dependency between initialization and propagation.

\begin{enumerate}
    \item \textbf{Phase I.} Optimize $\delta$ using only $\mathcal{L}_{\text{init}}$ on the initial frame, without propagation, to establish a corrupted initial memory anchor $\tilde{E}_1$. 

    \item \textbf{Phase II.} Extend training to full video sequence. The initial frame continues to be supervised by $\mathcal{L}_{\text{init}}$, while all subsequent frames are jointly penalized by $\mathcal{L}_{\text{prop}}^{(i)}$. 
\end{enumerate}

The overall objective for optimizing the UAP $\delta$ is formulated as:
\begin{equation}
\mathcal{L}_{\text{total}} =  \lambda \mathcal{L}_{\text{init}} + \sum_{i=2}^{T} \mathcal{L}_{\text{prop}}^{(i)},
\label{eq:ltotal}
\end{equation}
In both phases, $\delta$ is optimized by targeting memory and memory-attended features, avoiding supervision on predicted masks or other prompt-conditioned decoder outputs. This enables robust prompt-agnostic attacks. See Section~\ref{sec:prompt_agnostic} for in-depth analysis of the underlying mechanism.

%% file: supple.tex
\clearpage
\appendix
\setcounter{section}{0}
\renewcommand{\thesection}{\Alph{section}}
\renewcommand{\thesubsection}{\thesection.\arabic{subsection}}
\renewcommand{\theHsection}{supp.\Alph{section}}
\renewcommand{\theHsubsection}{supp.\Alph{section}.\arabic{subsection}}

\title{Attack-Inference Aligned Universal Adversarial Attacks on Video Object Segmentation}
\maketitle

\section*{Appendix}

\noindent The appendix is structured as follows:

\begin{enumerate}[label=\textbf{\Alph*.}, leftmargin=1.8em]
    \item Section~\ref{sec:benchmarks} provides a detailed description of all \textbf{benchmarks} used in evaluation, including both standard VOS benchmarks and promptable concept segmentation benchmarks.

    \item Section~\ref{sec:full_result} presents comprehensive \textbf{quantitative} comparisons between \methodname and prior attack baselines across extensive benchmarks and prompt types, along with analysis of temporal dynamics.

    \item Section~\ref{sec:algorithm} details the optimization \textbf{algorithm} of \methodname, including its two-phase curriculum for initialization-aligned memory corruption and propagation-aligned saliency flattening.

    \item Section~\ref{sec:memory_defenses} evaluates \methodname's performance under \textbf{memory-level defenses}, including random memory dropping and advanced memory-management strategies.

    \item Section~\ref{sec:robustness} further validates the robustness of \methodname through two \textbf{stress tests} on SAM3: 1) input prompt variations, including spatial transformations and changes in prompt quantity; and 2) stricter perturbation budgets, where $\epsilon$ is gradually reduced from standard $10/255$ all the way to $4/255$ to assess \methodname's efficacy.

    \item Section~\ref{sec:one-to-many} highlights the \textbf{one-to-many} attack capability of \methodname in language-based video segmentation, where a single language prompt corresponds to multiple targets and \methodname suppresses them all in one shot.
    
    \item Section~\ref{sec:image} further evaluates \methodname on \textbf{image segmentation benchmarks}. Specifically, we test its attack performance on ADE20K \cite{zhou2017scene} and RefCOCO \cite{mao2016generation, yu2016modeling} under point, box, and language prompts.
 
    \item Section~\ref{sec:vis} provides \textbf{qualitative} comparisons with prior attack baselines, visually demonstrating the robust attack performance of \methodname.

\end{enumerate}

\section{Benchmarks}
\label{sec:benchmarks}
We provide additional details on the benchmarks used to evaluate \methodname. 
Our evaluation covers both standard video object segmentation (VOS) benchmarks and promptable concept segmentation (PCS) benchmarks. The former focuses on tracking targets specified by masks, points, or bounding boxes. The latter evaluates language-based video segmentation, where targets are specified by natural language expressions and may correspond to one or multiple object instances. A detailed comparison of these benchmarks is given in Table \ref{tab:benchmark_summary}.

\subsection{Standard VOS Benchmarks}
We evaluate \methodname on four widely-adopted VOS benchmarks supporting different forms of target specification, including masks (semi-supervised VOS), points, and bounding boxes (interactive VOS). Details are as follows:

\paragraph{DAVIS17}\cite{pont20172017} A high-quality VOS benchmark with precise pixel-level annotations and natural language labels (e.g., \textit{pigs, dog, car-shadow}) for each video sequence.

\paragraph{YouTube-VOS 2019 (YTVOS19)}\cite{xu2018youtube} A large-scale VOS benchmark characterized by its diverse object categories and a significantly larger volume of video sequences compared to DAVIS17.

\paragraph{MOSEv1}\cite{ding2023mose} A challenging VOS benchmark focused on complex scenarios such as heavy occlusions, crowded environments, and frequent object disappearances and reappearances.

\paragraph{MOSEv2}\cite{ding2025mosev2} An extended version of MOSEv1 featuring more challenging cases, including adverse weather, low-light conditions, camouflaged objects, and scenarios requiring external knowledge.

\begin{table}[h]
\centering
\scriptsize
\caption{Summary of different benchmarks used in \methodname's evaluation.}
\label{tab:benchmark_summary}
\vspace{1mm}
\renewcommand{\arraystretch}{1}
\resizebox{\linewidth}{!}{
\begin{tabular}{l|c|c|l|c}
\toprule
Benchmark & Year & Task & Prompt Type & Videos (Val. split) \\
\midrule
DAVIS17~\cite{pont20172017} & 2017 & VOS & Mask / Point / Box & 30 \\
YTVOS19~\cite{xu2018youtube} & 2019 & VOS & Mask / Point / Box & 507 \\
MOSEv1~\cite{ding2023mose} & 2023 & VOS & Mask / Point / Box & 311 \\
MOSEv2~\cite{ding2025mosev2} & 2025 & VOS & Mask / Point / Box & 433 \\
\midrule
SA-Co/VEval--SA-V~\cite{carion2025sam} & 2025 & PCS & Language & 2k NPs \\
SA-Co/VEval--YT-Temporal-1B~\cite{carion2025sam} & 2025 & PCS & Language & 1.7k NPs \\
SA-Co/VEval--SmartGlasses~\cite{carion2025sam} & 2025 & PCS & Language & 2.4k NPs \\
\bottomrule
\end{tabular}
}
\vspace{-1mm}
\end{table}

\subsection{PCS Benchmarks}

\paragraph{SA-Co/VEval} \cite{carion2025sam} A large-scale language-based video segmentation benchmark introduced in SAM3. Given a noun phrase (NP) (e.g., \textit{"white car", "man in jeans"}), the model first detects and then tracks all corresponding targets throughout the entire sequence. This large-scale PCS benchmark comprises three subsets: SA-V (2k NPs), YT-Temporal-1B (1.7k NPs), and SmartGlasses (2.4k NPs).

\section{Full Results}
\label{sec:full_result}

Due to space constraints in the main paper, we provide full comparisons of different attack methods on SAM3 \cite{carion2025sam} across prompt modalities and benchmarks here. For a fair comparison, all baselines use \textbf{the same} training configurations as ours, as described in Section \ref{sec:implementation}.

\begin{table*}[!t]
\centering
\caption{Full comparison of \methodname and existing attack baselines on SAM3 \cite{carion2025sam} across VOS benchmarks and prompt modalities. All baselines use the same training configurations on SAM3 as our \methodname.}
\label{tab:supple_full_results}
\renewcommand{\arraystretch}{1}
\resizebox{\textwidth}{!}{
\begin{tabular}{l|ccc|ccc|ccc|ccc}
\toprule
\multirow{2}{*}{Methods} & \multicolumn{3}{c|}{DAVIS17 ($\mathcal{J}\&\mathcal{F}$$\downarrow$)} & \multicolumn{3}{c|}{YTVOS19 ($\mathcal{G}$$\downarrow$)} & \multicolumn{3}{c|}{MOSEv2 ($\mathcal{J}\&\dot{\mathcal{F}}$$\downarrow$)} & \multicolumn{3}{c}{MOSEv1 ($\mathcal{J}\&\mathcal{F}$$\downarrow$)} \\ \cmidrule{2-13}
& Mask & Point & Box & Mask & Point & Box & Mask & Point & Box & Mask & Point & Box \\ \midrule
Clean & 92.2 & 72.1 & 89.1 & 89.7 & 80.0 & 86.8 & 60.3 & 55.8 & 58.9 & 78.4 & 70.1 & 77.0 \\
PGD \cite{madry2017towards} & 65.5 & 21.9 & 58.8 & 55.9 & 27.6 & 49.9 & 35.0 & 28.9 & 34.3 & 46.4 & 33.0 & 43.9 \\
UAPGD \cite{deng2020universal} & 64.1 & 21.5 & 57.4 & 58.8 & 29.3 & 52.9 & 35.9 & 29.1 & 35.0 & 49.7 & 35.4 & 47.8 \\
SegPGD \cite{gu2022segpgd} & 59.4 & 17.5 & 52.4 & 50.3 & 22.3 & 43.8 & 32.6 & 27.0 & 32.3 & 42.0 & 28.0 & 39.8 \\
VOSPGD \cite{jiang2023exploring} & 61.2 & 19.8 & 54.5 & 49.5 & 21.8 & 42.5 & 35.0 & 28.1 & 33.5 & 44.5 & 30.5 & 42.1 \\
AttackSAM \cite{zhang2023attack} & 68.3 & 30.2 & 63.2 & 63.0 & 30.9 & 53.2 & 42.6 & 36.4 & 38.2 & 52.5 & 41.2 & 47.3 \\
DarkSAM \cite{zhou2024darksam} & 58.5 & 25.0 & 52.5 & 51.7 & 23.1 & 45.6 & 32.5 & 26.9 & 32.0 & 43.0 & 28.0 & 41.3 \\
UAP-SAM2 \cite{zhou2025vanish} & 55.4 & 23.9 & 47.4 & 46.3 & 22.0 & 35.8 & 30.4 & 25.5 & 29.1 & 36.4 & 22.5 & 31.3 \\ \midrule
\rowcolor{defaultColor} \textbf{\methodname (Ours)} & \ \ \textbf{8.4} & \ \ \textbf{7.1} & \ \ \textbf{7.6} & \textbf{14.0} & \textbf{10.3} & \textbf{12.7} & \textbf{24.6} & \textbf{19.9} & \textbf{23.5} & \textbf{21.1} & \textbf{18.3} & \textbf{20.2} \\ \bottomrule
\end{tabular}
}
\end{table*}

\begin{table*}[t]
\centering
\caption{Full comparison of attack temporal dynamics on SAM3. We omit MOSE because per-frame ground-truth masks are unavailable for its validation set.}
\label{tab:supple_temporal_analysis}
\renewcommand{\arraystretch}{1}
\small 
\setlength{\tabcolsep}{2.2pt} 
\begin{tabular}{l | ccc | ccc | ccc | ccc}
\toprule
\multirow{3}{*}{Method} & \multicolumn{6}{c|}{DAVIS17} & \multicolumn{6}{c}{YTVOS19} \\
\cmidrule(lr){2-7} \cmidrule(lr){8-13}
 & \multicolumn{3}{c|}{Mask} & \multicolumn{3}{c|}{Point} & \multicolumn{3}{c|}{Mask} & \multicolumn{3}{c}{Box} \\
\cmidrule(lr){2-4} \cmidrule(lr){5-7} \cmidrule(lr){8-10} \cmidrule(lr){11-13}
 & $Lat$$\downarrow$ & $Cov$$\uparrow$ & $Cov_{20}$$\uparrow$ 
 & $Lat$$\downarrow$ & $Cov$$\uparrow$ & $Cov_{20}$$\uparrow$ 
 & $Lat$$\downarrow$ & $Cov$$\uparrow$ & $Cov_{20}$$\uparrow$ 
 & $Lat$$\downarrow$ & $Cov$$\uparrow$ & $Cov_{20}$$\uparrow$ \\
\midrule
PGD~\cite{madry2017towards}
& 61.3 & 28.3 & 34.0 & 12.3 & 82.5 & 82.8 & 61.2 & 31.7 & 38.0 & 52.7 & 39.4 & 47.2 \\
UAPGD~\cite{deng2020universal} 
& 59.0 & 27.7 & 35.7 & 15.1 & 78.1 & 79.1 & 61.2 & 29.1 & 33.7 & 51.4 & 36.6 & 40.8 \\
SegPGD~\cite{gu2022segpgd} 
& 53.7 & 36.5 & 45.7 & 10.0 & 86.4 & 87.5 & 53.9 & 38.6 & 46.4 & 46.6 & 46.6 & 55.3 \\
AttackSAM~\cite{zhang2023attack} 
& 60.4 & 29.2 & 36.1 & 18.3 & 75.5 & 76.7 & 62.6 & 29.6 & 35.7 & 47.3 & 44.0 & 51.1 \\
DarkSAM~\cite{zhou2024darksam} 
& 58.3 & 33.5 & 41.6 & 11.0 & 84.6 & 85.0 & 56.1 & 36.6 & 43.8 & 46.6 & 44.8 & 52.6 \\
UAP-SAM2~\cite{zhou2025vanish} 
& 52.5 & 35.9 & 43.6 & 10.4 & 85.1 & 85.2 & 50.9 & 41.1 & 47.4 & 36.1 & 56.6 & 63.5 \\
\midrule
\rowcolor{defaultColor} 
\textbf{\methodname (Ours)} 
& \ \ \textbf{5.5} & \textbf{89.2} & \textbf{88.1} 
& \ \ \textbf{2.2} & \textbf{90.3} & \textbf{88.6} 
& \ \ \textbf{8.6} & \textbf{85.1} & \textbf{84.6} 
& \ \ \textbf{7.0} & \textbf{86.8} & \textbf{85.9} \\
\bottomrule
\end{tabular}
\end{table*}

As shown in Table~\ref{tab:supple_full_results}, \methodname consistently surpasses existing attack baselines across all VOS benchmarks and prompt types, including mask, point, and box prompts. 
For example, on DAVIS17-val with point prompts, \methodname achieves 7.1 $\mathcal{J}\&\mathcal{F}$, outperforming the previous SOTA UAP-SAM2~\cite{zhou2025vanish} of 23.9, even though the latter is specifically optimized for point-based attacks. On more challenging benchmarks such as MOSEv2 and MOSEv1, \methodname also achieves the best performance across all prompt types. Notably, on MOSEv1 with mask prompts, \methodname outperforms previous SOTA by 15.3 $\mathcal{J}\&\mathcal{F}$.

Also, as discussed in Section~\ref{sec:metrics} of the main paper, standard VOS metrics do not capture the temporal dynamics of an attack. Therefore, we further compare \methodname with the baselines in terms of attack onset ($Lat$), persistence ($Cov$), and mid-video persistence ($Cov_k$). As shown in Table~\ref{tab:supple_temporal_analysis}, \methodname demonstrates a clear advantage across all temporal metrics: it takes effect much faster, achieves long-lasting suppression, and exhibits robust online attack capability, driven by its propagation-aligned training on memory-based propagation.

For language-based segmentation, \methodname achieves the strongest attack performance across all three subsets of SA-Co/VEval, as shown in Table \ref{tab:supple_language}. Notably, it reduces cgF1 and pHOTA on YT-Temporal-1B to 0.3 and 9.1, respectively, significantly surpassing the previous SOTA of UAP-SAM2 (3.5 and 26.2, respectively). We further visualize its one-to-many attack capability using language prompts in Section \ref{sec:one-to-many}.

\begin{table*}[!t]
\centering
\caption{Full comparison of \methodname and existing attack baselines on SAM3 across \textbf{language-based} SA-Co/VEval benchmarks~\cite{carion2025sam}. Lower values are better for all metrics.}
\renewcommand{\arraystretch}{1}
\label{tab:supple_language}
\resizebox{\textwidth}{!}{
\begin{tabular}{l|cccc|cccc|cccc}
\toprule
\multirow{2}{*}{Methods} & \multicolumn{4}{c|}{SA-V} & \multicolumn{4}{c|}{YT-Temporal-1B} & \multicolumn{4}{c}{SmartGlasses} \\ \cmidrule{2-13}
& cgF1 & pHOTA & pDetA & pAssA & cgF1 & pHOTA & pDetA & pAssA & cgF1 & pHOTA & pDetA & pAssA \\ \midrule
Clean & 29.3 & 60.7 & 44.7 & 83.2 & 50.2 & 70.5 & 60.5 & 82.7 & 33.5 & 60.2 & 46.2 & 79.3 \\
PGD~\cite{madry2017towards} & \ \ 5.2 & 33.5 & 15.1 & 74.7 & 13.2 & 43.7 & 27.8 & 69.4 & 14.7 & 42.0 & 26.7 & 66.9 \\
UAPGD~\cite{deng2020universal} & \ \ 8.1 & 38.1 & 20.3 & 72.5 & 17.4 & 45.1 & 30.8 & 66.9 & 16.5 & 44.2 & 30.0 & 65.9 \\
SegPGD~\cite{gu2022segpgd} & \ \ 4.0 & 31.7 & 13.5 & 75.8 & 11.5 & 40.2 & 24.5 & 66.6 & 13.6 & 41.8 & 26.4 & 67.0 \\
UAD~\cite{lu2024unsegment} & \ \ 2.4 & 24.9 & \ \ 8.5 & 74.0 & \ \ 7.8 & 31.7 & 15.0 & 67.5 & 11.1 & 36.6 & 21.0 & 64.4 \\
DarkSAM~\cite{zhou2024darksam} & \ \ 4.0 & 30.5 & 12.6 & 75.0 & 10.7 & 38.8 & 22.7 & 67.0 & 12.4 & 38.8 & 23.0 & 65.9 \\
UAP-SAM2~\cite{zhou2025vanish} & \ \ 1.7 & 23.9 & \ \ 7.8 & 74.6 & \ \ 3.5 & 26.2 & 11.1 & 62.4 & \ \ 9.1 & 36.3 & 21.4 & 62.4 \\ \midrule
\rowcolor{defaultColor} \textbf{\methodname (Ours)} & \textbf{\ \ 0.1} & \textbf{\ \ 8.5} & \textbf{\ \ 1.2} & \textbf{64.3} & \textbf{\ \ 0.3} & \textbf{\ \ 9.1} & \textbf{\ \ 1.7} & \textbf{49.9} & \textbf{\ \ 1.9} & \textbf{18.4} & \textbf{\ \ 6.4} & \textbf{53.4} \\ \bottomrule
\end{tabular}
}
\end{table*}

\section{Algorithm}
\label{sec:algorithm}

Algorithm~\ref{alg:algorithm} summarizes the optimization process of \methodname, which aligns with the two-stage inference paradigm of modern VOS models through a two-phase curriculum.

In Phase I, we optimize the UAP $\delta$ to corrupt the initial memory anchor $\tilde{E}_1$ before it is permanently stored in the memory bank. In Phase II, we jointly optimize the same $\delta$ to corrupt the initial memory anchor and flatten the spatial saliency of memory-attended feature maps $\tilde{A}_i$, disrupting both initialization and propagation stages.
 
\section{Memory-Level Defenses}
\label{sec:memory_defenses}
Extending Section~\ref{sec:defense}, we evaluate two additional memory-level defenses on DAVIS17 using SAM2.1-B+ \cite{ravi2024sam} and its extensions with more advanced memory management. We first apply random memory dropping, discarding 25\%, 50\%, or 75\% of the non-conditioning memory entries at each frame during propagation. We then evaluate different SAM2 extensions, including SAM2Long \cite{ding2025sam2long}, SAMURAI \cite{yang2024samurai}, and DAM4SAM \cite{videnovic2025distractor}, which incorporate their respective memory-management strategies. As shown in Table~\ref{tab:memory_defenses}, \methodname remains highly effective across different memory-level defenses.
\vspace{-1em}

\begin{table}[h]
\centering
\caption{Attack performance ($\mathcal{J}\&\mathcal{F}$) under memory-level defenses on SAM2.1-B+.}
\label{tab:memory_defenses}
\vspace{1mm}
\small
\setlength{\tabcolsep}{5pt}
\renewcommand{\arraystretch}{1.12}
\begin{tabular}{@{}llccc@{}}
\toprule
\textbf{Defense} & \textbf{Memory Strategy}
& \textbf{Mask}$\downarrow$
& \textbf{Point}$\downarrow$
& \textbf{Box}$\downarrow$ \\
\midrule
Random dropping (25\%) & Remove non-conditioning entries
& 15.8 & 7.6 & 11.6 \\
Random dropping (50\%) & Remove non-conditioning entries
& 16.4 & 7.8 & 11.9 \\
Random dropping (75\%) & Remove non-conditioning entries
& 18.2 & 8.4 & 12.7 \\
\midrule
SAM2Long (SAM2.1-B+) & Memory-tree search
& 15.9 & 8.6 & 11.5 \\
SAMURAI (SAM2.1-B+) & Motion-aware memory selection
& 20.7 & 7.8 & 9.0 \\
DAM4SAM (SAM2.1-B+) & Distractor-aware memory selection
& 15.8 & 7.3 & 14.1 \\
\bottomrule
\end{tabular}
\end{table}

\section{Further Robustness Analysis}
\label{sec:robustness}

\subsection{Prompt Variations}

We adopt the standard practice for evaluating point and box prompts in the main paper, following UAP-SAM2 \cite{zhou2025vanish} and DarkSAM \cite{zhou2024darksam}, which uses the center point and tight bounding box derived from the target mask, respectively. To further evaluate the robustness of \methodname, we test if prompt variations can successfully bypass the perturbation. Specifically, we evaluate (1) random shifts in point positions, (2) variations in point count, and (3) random translation and scaling of bounding boxes.

As shown in Table \ref{tab:robustness}, \methodname maintains consistent attack performance across prompt variations, demonstrating strong robustness to changes in \textbf{prompt position, translation, and count}. These results further indicate that \methodname generalizes across diverse prompt patterns rather than relying on a specific prompt distribution.

\begin{algorithm}[!t]
\caption{\methodname: Attack-Inference Aligned Universal Adversarial Perturbation on Video Object Segmentation.}
\label{alg:algorithm}
\small
\begin{algorithmic}[1]
\Require A training video $\mathcal{V}$, VOS model $\Phi=(\Phi_{\mathrm{init}},\Phi_{\mathrm{prop}})$,
memory encoder $\mathcal{E}_{\mathrm{mem}}$, perturbation bound $\epsilon$, step size $\eta$,
phase epochs $K_{\mathrm{init}},K_{\mathrm{prop}}$, trade-off $\lambda$
\Ensure Universal adversarial perturbation $\delta$
\Statex \textit{Mask outputs below denote soft values before binarization.}
\State Initialize $\delta \in \mathbb{R}^{H \times W \times 3}$ and project
$\delta \gets \Pi_{\|\delta\|_{\infty}\leq \epsilon}(\delta)$
\Statex \textbf{Phase I: Initialization-only optimization}
\For{$e=1$ to $K_{\mathrm{init}}$}
    \For{video $\mathcal{V}=\{I_1,\ldots,I_T\}$ with prompt $P$}
        \State $M_1^{\mathrm{soft}} \gets \Phi_{\mathrm{init}}(I_1,P)$
        \State $E_1 \gets \mathcal{E}_{\mathrm{mem}}(I_1+\delta,M_1^{\mathrm{soft}})$,
        \quad $E_1^{\emptyset} \gets \mathcal{E}_{\mathrm{mem}}(I_1+\delta,\mathbf{0})$
        \Comment{$E_1 \in \mathbb{R}^{h \times w \times c}$}
        \State $\widetilde{M_1}^{\mathrm{soft}} \gets \Phi_{\mathrm{init}}(I_1+\delta,P)$
        \State $\tilde{E}_1 \gets \mathcal{E}_{\mathrm{mem}}(I_1+\delta,\widetilde{M_1}^{\mathrm{soft}})$
        \State $\mathcal{L}_{\mathrm{init}} \gets
        \frac{\|\tilde{E}_1-E_1^{\emptyset}\|_2^2}{hwc}
        +\exp\!\left(-\frac{10\|\tilde{E}_1-E_1\|_2^2}{hwc}\right)$
        \State $\delta \gets
        \Pi_{\|\delta\|_{\infty}\leq \epsilon}
        \left(\delta-\eta\nabla_{\delta}\mathcal{L}_{\mathrm{init}}\right)$
    \EndFor
\EndFor
\Statex \textbf{Phase II: Joint initialization and propagation optimization}
\For{$e=1$ to $K_{\mathrm{prop}}$}
    \For{video $\mathcal{V}=\{I_1,\ldots,I_T\}$ with prompt $P$}
        \State $M_1^{\mathrm{soft}} \gets \Phi_{\mathrm{init}}(I_1,P)$
        \State $E_1 \gets \mathcal{E}_{\mathrm{mem}}(I_1+\delta,M_1^{\mathrm{soft}})$,
        \quad $E_1^{\emptyset} \gets \mathcal{E}_{\mathrm{mem}}(I_1+\delta,\mathbf{0})$
        \State $\widetilde{M_1}^{\mathrm{soft}} \gets \Phi_{\mathrm{init}}(I_1+\delta,P)$
        \State $\tilde{E}_1 \gets \mathcal{E}_{\mathrm{mem}}(I_1+\delta,\widetilde{M_1}^{\mathrm{soft}})$
        \State $\mathcal{L}_{\mathrm{init}} \gets
        \frac{\|\tilde{E}_1-E_1^{\emptyset}\|_2^2}{hwc}
        +\exp\!\left(-\frac{10\|\tilde{E}_1-E_1\|_2^2}{hwc}\right)$
        \State $\mathcal{B} \gets \{\tilde{E}_1\}$
        \State $\mathcal{L}_{\mathrm{prop}} \gets 0$
        \For{$i=2$ to $T$}
            \State $F_i \gets \mathcal{E}_{\mathrm{img}}(I_i+\delta)$
            \State $\mathbf{B}_i \gets \operatorname{Read}(\mathcal{B})$
            \State $\tilde{A}_i \gets
            \operatorname{softmax}\!\left(F_i \mathbf{B}_i^{\top}\right)
            \mathbf{B}_i + F_i$
            \State $\tilde{\mu}_i \gets
            \frac{1}{hw}\sum_{x=1}^{h}\sum_{y=1}^{w}
            \tilde{A}_i(x,y,:)$
            \State $\mathcal{L}_{\mathrm{prop}} \gets
            \mathcal{L}_{\mathrm{prop}}+
            \frac{1}{hwc}\sum_{x=1}^{h}\sum_{y=1}^{w}
            \left\|\tilde{A}_i(x,y,:)-\tilde{\mu}_i\right\|_2^2$
            \State $\widetilde{M_i}^{\mathrm{soft}} \gets \operatorname{Decoder}(\tilde{A}_i)$
            \Comment{no prompt input during propagation}
            \State $\tilde{E}_i \gets \mathcal{E}_{\mathrm{mem}}(I_i+\delta,\widetilde{M_i}^{\mathrm{soft}})$
            \State $\mathcal{B} \gets \operatorname{FIFO}(\mathcal{B},\tilde{E}_i)$
            \Comment{always retain the initial anchor $\tilde{E}_1$}
        \EndFor
        \State $\mathcal{L}_{\mathrm{total}}
        \gets \lambda\mathcal{L}_{\mathrm{init}}+\mathcal{L}_{\mathrm{prop}}$
        \State $\delta \gets
        \Pi_{\|\delta\|_{\infty}\leq \epsilon}
        \left(\delta-\eta\nabla_{\delta}\mathcal{L}_{\mathrm{total}}\right)$
    \EndFor
\EndFor
\State \Return $\delta$
\end{algorithmic}
\end{algorithm}

\subsection{Stricter Perturbation Bound}
To evaluate the robustness of \methodname under stricter perturbation bound, we reduce $\epsilon$ from the standard $10/255$ \cite{moosavi2017universal, deng2020universal, naseer2020self, zhou2024darksam, zhou2025vanish} to $8/255$, $6/255$, and $4/255$, respectively.

As shown in Table~\ref{tab:epsilon_ablation}, \methodname remains effective under smaller perturbation budgets. Remarkably, even at the strictest bound of $4/255$, its performance remains \textbf{on par with} the previous SOTA of UAP-SAM2 \cite{zhou2025vanish}, which uses a much larger perturbation budget of $10/255$. For example, on mask-based YTVOS19, \methodname ($4/255$) achieves 41.0 $\mathcal{G}$ compared to 46.3 for UAP-SAM2 ($10/255$), while outperforming the latter on the challenging benchmarks of MOSEv2 and MOSEv1. Furthermore, at $\epsilon = 6/255$, \methodname already establishes new SOTA results across different benchmarks compared to the previous SOTA.

\begin{table*}[!t]
\caption{Stress test of prompt variations, evaluated on DAVIS17 across various point positions (left), number of points (middle), and bounding box transformations (right).}
\label{tab:robustness}
\renewcommand{\arraystretch}{1}
\centering
\small
\setlength{\tabcolsep}{3.5pt} 
\begin{minipage}{0.32\textwidth}
\centering
\begin{tabular}{l|ccc}
\toprule
\multirow{2}{*}{Position} & \multicolumn{3}{c}{Point Position} \\ \cmidrule{2-4}
 & $\mathcal{J}\&\mathcal{F}\downarrow$ & $\mathcal{J}\downarrow$ & $\mathcal{F}\downarrow$ \\ \midrule
Center & 7.1 & 7.1 & 7.0 \\
Top-left         & 6.4 & 6.2 & 6.5 \\
Top-right        & 6.3 & 6.2 & 6.3 \\
Bottom-left      & 6.5 & 6.3 & 6.6 \\
Bottom-right     & 5.3 & 5.1 & 5.4 \\ \bottomrule
\end{tabular}
\end{minipage}
\hfill \hfill \hfill
\begin{minipage}{0.32\textwidth}
\centering
\begin{tabular}{c|ccc}
\toprule
\multirow{2}{*}{Num.} & \multicolumn{3}{c}{Point Number} \\ \cmidrule{2-4}
 & $\mathcal{J}\&\mathcal{F}\downarrow$ & $\mathcal{J}\downarrow$ & $\mathcal{F}\downarrow$ \\ \midrule
1 & 7.1 & 7.1 & 7.0 \\
2           & 7.5 & 7.5 & 7.5 \\
4           & 7.4 & 7.3 & 7.4 \\
8           & 7.2 & 7.2 & 7.2 \\
16          & 6.9 & 6.8 & 6.9 \\ \bottomrule
\end{tabular}
\end{minipage}
\hfill
\begin{minipage}{0.32\textwidth}
\centering
\begin{tabular}{l|ccc}
\toprule
\multirow{2}{*}{Position} & \multicolumn{3}{c}{Box Position} \\ \cmidrule{2-4}
 & $\mathcal{J}\&\mathcal{F}\downarrow$ & $\mathcal{J}\downarrow$ & $\mathcal{F}\downarrow$ \\ \midrule
Center & 7.6 & 7.5 & 7.7 \\
Shift-left      & 7.7 & 7.4 & 7.9 \\
Shift-right     & 7.6 & 7.4 & 7.8 \\
Scale-up        & 7.7 & 7.5 & 7.8 \\
Scale-down      & 7.3 & 7.0 & 7.5 \\ \bottomrule
\end{tabular}
\end{minipage}
\end{table*}

\begin{table*}[!t]
\centering
\caption{Stress test of lowering the perturbation bound $\epsilon$. We evaluate \methodname by gradually reducing $\epsilon$ from the standard $10/255$ all the way to $4/255$.}
\label{tab:epsilon_ablation}
\renewcommand{\arraystretch}{1.1}
\resizebox{\textwidth}{!}{
\begin{tabular}{l|ccc|ccc|ccc|ccc}
\toprule
\multirow{2}{*}{Method / Budget $\epsilon$} & \multicolumn{3}{c|}{DAVIS17 ($\mathcal{J}\&\mathcal{F}$$\downarrow$)} & \multicolumn{3}{c|}{YTVOS19 ($\mathcal{G}$$\downarrow$)} & \multicolumn{3}{c|}{MOSEv2 ($\mathcal{J}\&\dot{\mathcal{F}}$$\downarrow$)} & \multicolumn{3}{c}{MOSEv1 ($\mathcal{J}\&\mathcal{F}$$\downarrow$)} \\ \cmidrule{2-13}
& Mask & Point & Box & Mask & Point & Box & Mask & Point & Box & Mask & Point & Box \\ \midrule
Clean & 92.2 & 72.1 & 89.1 & 89.7 & 80.0 & 86.8 & 60.3 & 55.8 & 58.9 & 78.4 & 70.1 & 77.0 \\ \midrule
UAP-SAM2 ($10/255$) \cite{zhou2025vanish} & 55.4 & 23.9 & 47.4 & 46.3 & 22.0 &  35.8 & 30.4 & 25.5 & 29.1 &  36.4 & 22.5 & 31.3 \\ \midrule
\rowcolor{defaultColor}  \textbf{\methodname ($10/255$)} & \textbf{\ \ 8.4} & \textbf{\ \ 7.1} & \textbf{\ \ 7.6} & \textbf{14.0} & \textbf{10.3} & \textbf{12.7} & \textbf{24.6} & \textbf{19.9} & \textbf{23.5} & \textbf{21.1} & \textbf{18.3} & \textbf{20.2} \\
\methodname ($8/255$) & 12.3 & \ \ 9.1 & 11.4 & 17.6 & 13.2 & 16.8 & 23.1 & 18.4 & 22.7 & 24.8 & 21.3 & 24.1 \\
\methodname ($6/255$) & 18.1 & 14.5 & 16.4 & 23.7 & 16.9 & 21.0 & 29.7 & 25.1 & 28.1 & 32.4 & 26.1 & 31.0 \\
\methodname ($4/255$) & 55.6 & 41.8 & 51.7 & 41.0 & 29.9 & 37.4 & 21.0 & 17.5 & 20.0 & 30.3 & 22.4 & 30.3 \\\bottomrule
\end{tabular}
}
\vspace{-0.8em}
\end{table*}

\begin{figure*}[!t]
    \centering
    \includegraphics[width=1.0\textwidth]{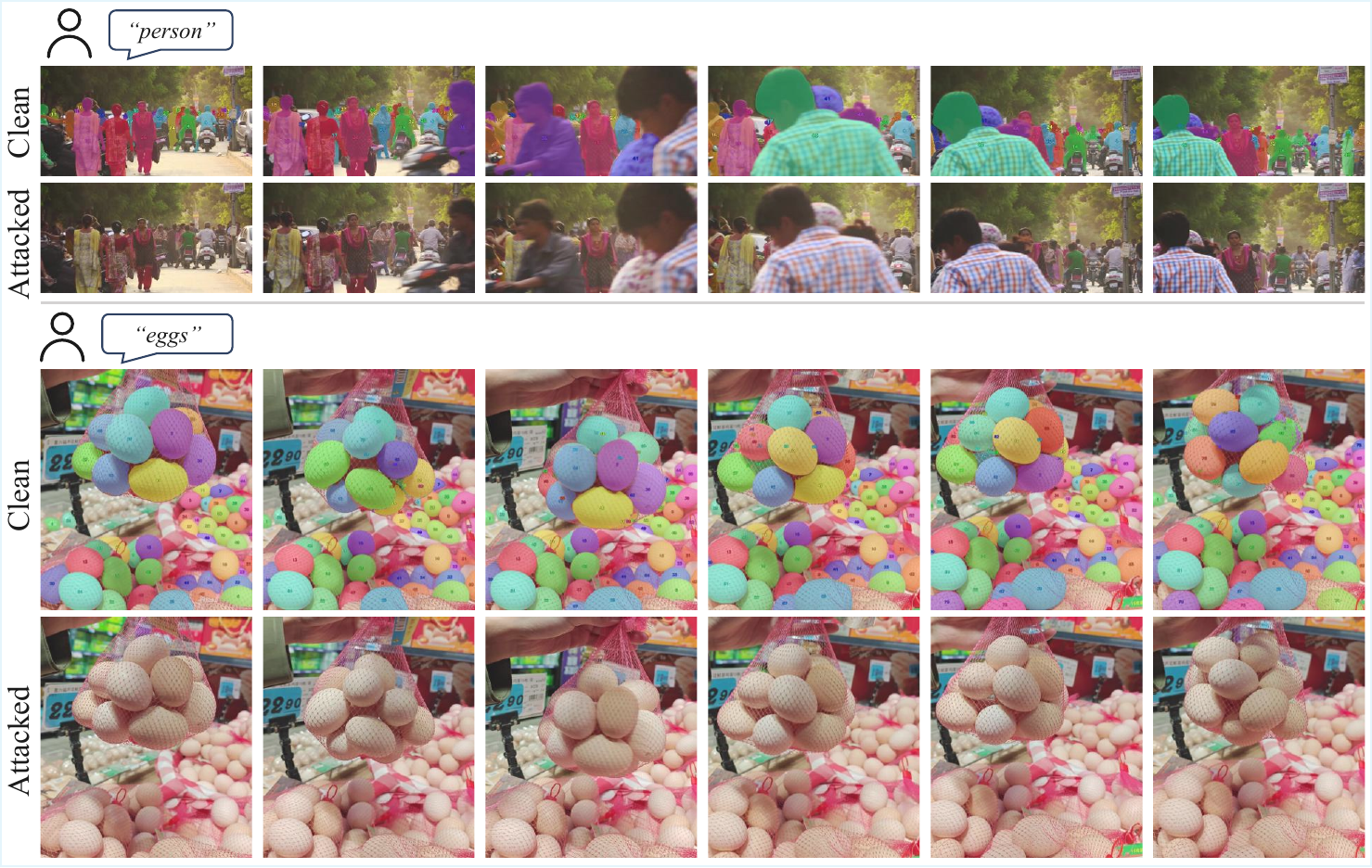}
    \caption{Multi-target video suppression by \methodname. It simultaneously eliminates multiple targets specified by a single language prompt, including people in a crowd (top) and eggs (bottom).}
    \label{fig:one_to_many}
\end{figure*}

\section{One-to-Many: Multi-Object Video Suppression}
\label{sec:one-to-many}
\vspace{-0.5em}
We further evaluate \methodname in one-to-many scenarios, where a single language prompt specifies multiple objects.

As shown in Figure~\ref{fig:one_to_many}, \methodname disrupts the segmentation of all specified targets, including people in a crowd (top) and eggs (bottom), causing their predicted masks to disappear immediately. This shows that \methodname naturally extends from individual objects to target groups.

\section{Attack on images}
\label{sec:image}

Following \cite{zhou2024darksam}, we further evaluate the attack performance of \methodname on image segmentation. Specifically, we conduct experiments on SAM3 \cite{carion2025sam} using the validation splits of two image segmentation benchmarks: ADE20K \cite{zhou2017scene} and RefCOCO \cite{mao2016generation, yu2016modeling}. For ADE20K, a standard image segmentation benchmark, we select the first annotated target in each image and use either point or box prompts for interactive segmentation. For RefCOCO, a referring image segmentation benchmark, we directly use the provided language expression as the language prompt.

As shown in Table~\ref{tab:image_seg}, \methodname consistently achieves SOTA attack performance on image segmentation benchmarks, outperforming prior baselines across all settings, including point, box, and language. For instance, on box-based ADE20K, \methodname reduces mIoU to 15.8, outperforming the previous SOTA of 61.9. Similarly, on language-based RefCOCO, it achieves an mIoU of 12.2, compared with the previous SOTA of 32.1. These results demonstrate that \methodname generalizes effectively to both image and video segmentation attacks.

\begin{table}[!t]
\centering
\caption{Comparison of \methodname and baselines on image segmentation benchmarks.}
\vspace{0.5em}
\scriptsize
\label{tab:image_seg}
\renewcommand{\arraystretch}{1}
\resizebox{\linewidth}{!}{
\begin{tabular}{lccccccc}
\toprule
\multirow{2}{*}{Settings}
& \multicolumn{3}{c}{ADE20K (mIoU)}
& \multicolumn{4}{c}{RefCOCO} \\
\cmidrule(lr){2-4} \cmidrule(lr){5-8}
& Box & 3-Points & 1-Point
& mIoU & Pr@0.5 & Pr@0.7 & Pr@0.9 \\
\midrule
Clean   & 81.4 & 77.3 & 46.7 & 60.9 & 67.3 & 61.5 & 30.4 \\
PGD \cite{madry2017towards}  & 76.4 & 61.3 & 25.5 & 45.2 & 49.2 & 40.8 & 15.9 \\
UAPGD \cite{deng2020universal}  & 73.7 & 65.5 & 23.3 & 50.4 & 55.7 & 46.9 & 18.2 \\
SEGPGD \cite{gu2022segpgd} & 74.7 & 56.9 & 16.0 & 44.0 & 47.6 & 39.5 & 15.3 \\
DarkSAM \cite{zhou2024darksam} & 75.1 & 59.3 & 18.7 & 44.0 & 47.9 & 39.1 & 14.7 \\
UAP-SAM2 \cite{zhou2025vanish} & 61.9 & 37.2 & \ \ 8.5 & 32.1 & 35.0 & 27.3 & \ \ 8.3 \\
\rowcolor{defaultColor} \textbf{\methodname (Ours)} & \textbf{15.8} & \textbf{11.9} & \textbf{\ \ 7.3} & \textbf{12.2} & \textbf{12.1} & \textbf{\ \ 9.4} & \textbf{\ \ 2.7} \\
\bottomrule
\end{tabular}
}
\end{table}

\section{Qualitative Comparisons}
\label{sec:vis}

\begin{figure*}[!h]
    \centering
    \includegraphics[width=1.0\textwidth]{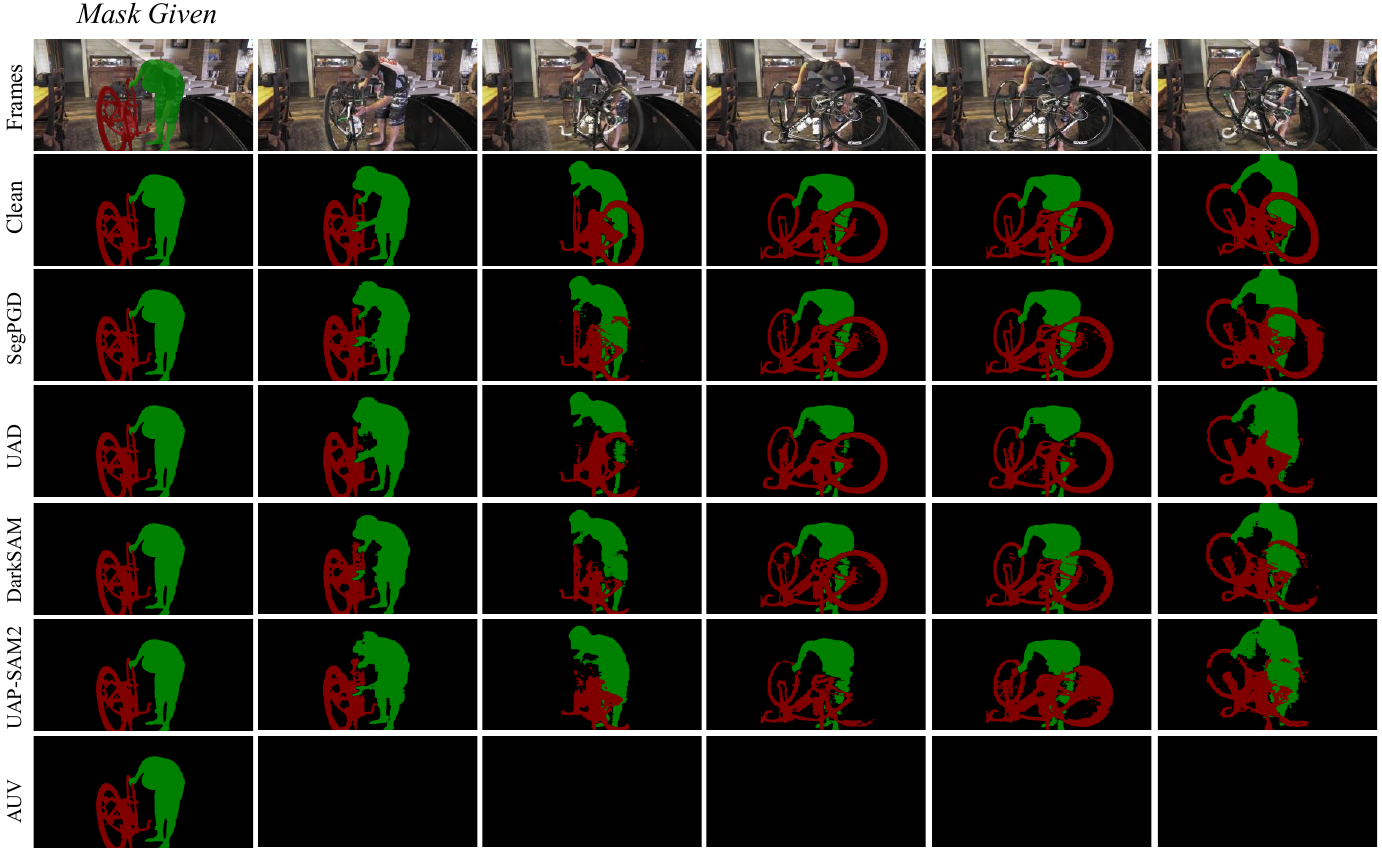}
    \caption{Qualitative comparison under mask prompts.}
    \label{fig:mask_vis}
\end{figure*}

\begin{figure*}[!t]
    \centering
    \includegraphics[width=1.0\textwidth]{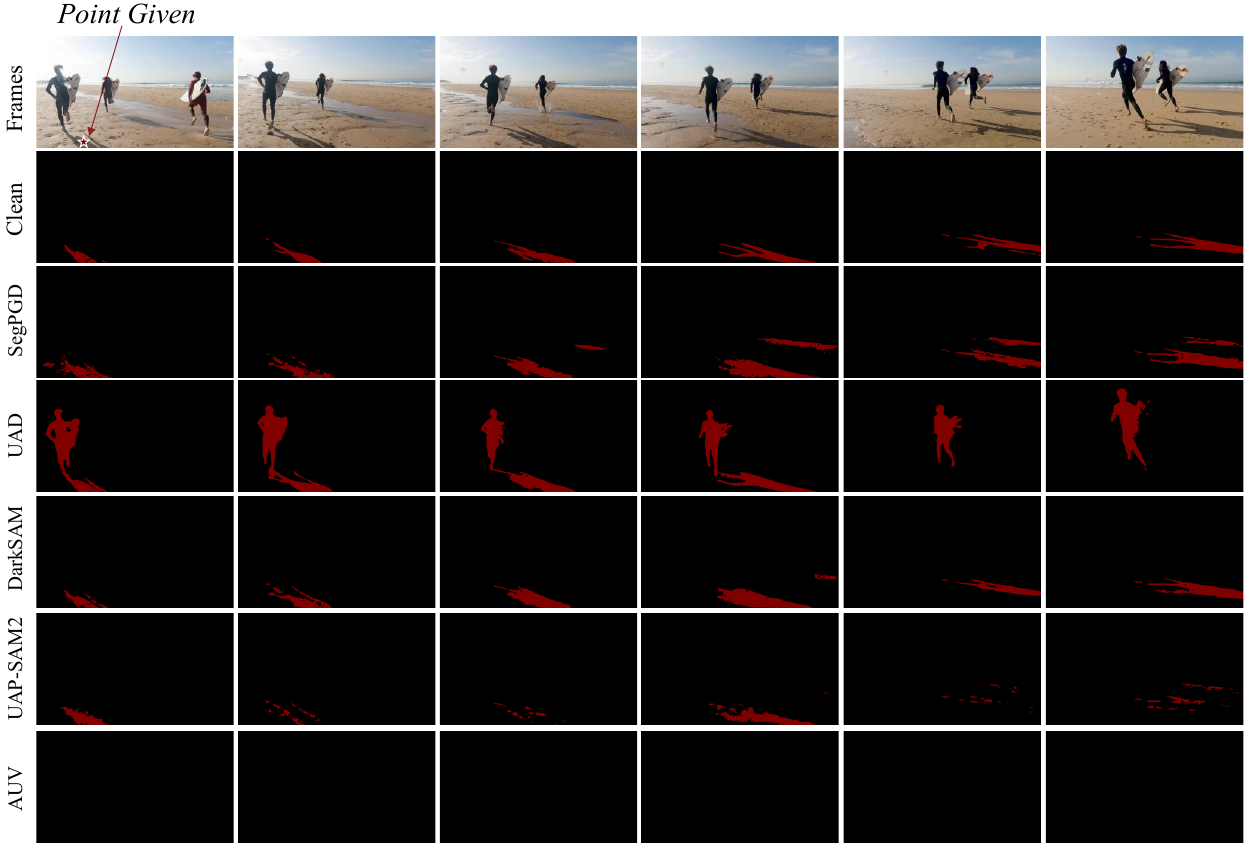}
    \caption{Qualitative comparison under point prompts.}
    \label{fig:point_vis}
\end{figure*}

\begin{figure*}[!t]
    \centering
    \includegraphics[width=1.0\textwidth]{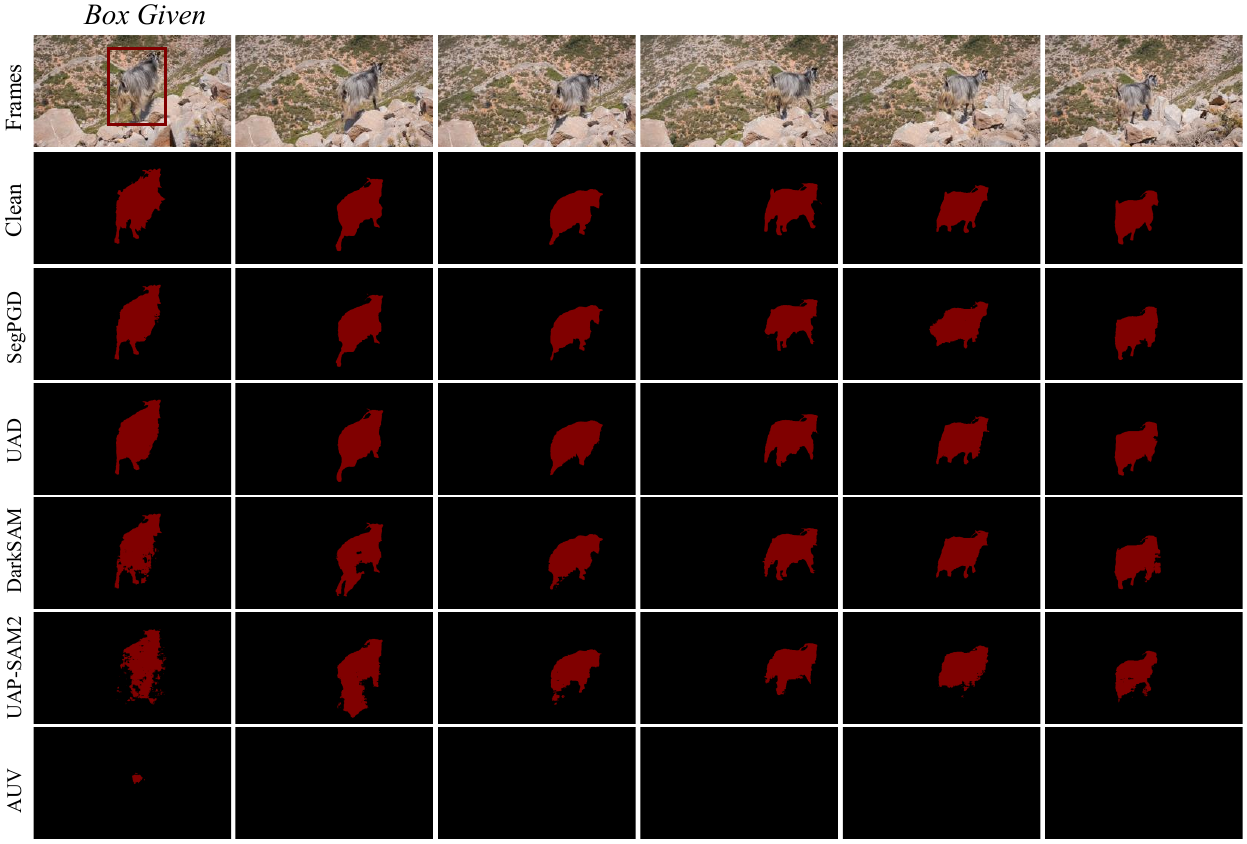}
    \caption{Qualitative comparison under box prompts.}
    \label{fig:box_vis}
\end{figure*}

Finally, we provide qualitative comparisons between \methodname and previous baselines, including SegPGD \cite{gu2022segpgd}, UAD \cite{lu2024unsegment}, DarkSAM \cite{zhou2024darksam}, and UAP-SAM2 \cite{zhou2025vanish}. Results on clean videos are also provided for reference. We compare the attack performance across different prompt modalities: mask, point, and box, which are used to specify the target on the initial frame. The visual comparisons are shown in Figure \ref{fig:mask_vis}, \ref{fig:point_vis}, and \ref{fig:box_vis}, respectively.

As shown, \methodname consistently outperforms all baselines across different prompt modalities. Specifically, it achieves the fastest onset and the most sustained attack coverage, maintaining strong effect throughout the entire video sequence regardless of the prompts used.

